\input{Def}
\documentclass[10pt,journal,compsoc]{IEEEtran}

\usepackage{times}
\usepackage{epsfig}
\usepackage{graphicx}
\usepackage{amsmath}
\usepackage{amssymb}

\usepackage{rotating}
\usepackage{multirow}
\usepackage{setspace}
\usepackage{algorithm2e}
\usepackage{fancyhdr} % to change header and footers
 \usepackage{color}
\usepackage{array}
\usepackage{subfigure}
\usepackage{soul}
\usepackage{verbatim} 
\usepackage{paralist}
\usepackage{xspace}
\usepackage{booktabs}
\usepackage{enumitem}
\usepackage{hyperref}

\ifCLASSOPTIONcompsoc
  \usepackage[nocompress]{cite}
\else
  \usepackage{cite}
\fi
\ifCLASSINFOpdf
\else
\fi

\begin{document}
\title{Dense Relational Image Captioning\\via Multi-task Triple-Stream Networks}

\author{Dong-Jin Kim,
        Tae-Hyun Oh,
        Jinsoo Choi, and
        In So Kweon,~\IEEEmembership{Members,~IEEE,}% <-this % stops a space
\IEEEcompsocitemizethanks{\IEEEcompsocthanksitem D. Kim, J. Choi, and I. S. Kweon are with the School of Electrical Engineering, KAIST, Daejeon, Republic of Korea.\protect\\
E-mail: \{djnjusa,jinsc37,iskweon77\}@kaist.ac.kr
\IEEEcompsocthanksitem T.-H. Oh is with the Dept. of Electrical Engineering and Graduate School of AI (GSAI), POSTECH, Pohang, Republic of Korea.
% the Computer Science and Artificial Intelligence Laboratory, Massachusetts Institute of Technology, Cambridge, MA, USA.
\protect\\
E-mail: 
% taehyun@csail.mit.edu
taehyun@postech.ac.kr
\IEEEcompsocthanksitem T.-H. Oh and I. S. Kweon are the co-corresponding authors of this work.
}% <-this % stops an unwanted space
\thanks{}}


%\markboth{IEEE TRANSACTION ON PATTERN ANALYSIS AND MACHINE INTELLIGENCE,~Vol.~14, No.~8, August~2015}%
%{Shell \MakeLowercase{\textit{et al.}}: Bare Demo of IEEEtran.cls for Computer Society Journals}

\IEEEtitleabstractindextext{%
\begin{abstract}
%\js{Our goal in this work is to train an image captioning model that generates more dense and informative captions.}{}
We introduce {\emph{dense relational captioning,}} a novel image captioning task which aims to generate multiple captions with respect to relational information between objects in a visual scene.
Relational captioning provides explicit descriptions for each relationship between object combinations. 
This framework is advantageous in both diversity and amount of information, leading to a {comprehensive} image understanding based on relationships, 
\eg, \emph{relational proposal generation}.
{For relational understanding between objects, the part-of-speech (POS; \ie, subject-object-predicate categories) can be a valuable prior information}
to guide the causal sequence of words in a caption.
% \js{We leverage the POS as a prior to guide the correct sequence of words in a caption.}
{We enforce our framework to learn not only to generate captions but also to understand the POS of each word.}
To this end, we propose {the} multi-task triple-stream network (MTTSNet) which consists of three recurrent units {responsible} for {each} POS {which is trained by} jointly {predicting the correct captions and POS for each word.} 
In addition, we found that the performance of MTTSNet can be improved by modulating the object embeddings with an explicit relational module.
We demonstrate {that our proposed model can generate} more diverse and richer {captions, via extensive experimental analysis on large scale datasets and several metrics.} %\js{generated by the proposed model against several baselines and competing methods.}{}
% {We additionally provide an ablation study, and applications on holistic image captioning, scene graph generation, and retrieval tasks.}
Then, we present 
% extend the analysis to an ablation study, 
applications of our framework to holistic image captioning, scene graph generation, and retrieval tasks.
%, so that we can understand the algorithmic behavior of our proposed method in various scenarios.
%Code has been made available at: \textcolor{magenta}{\url{https://github.com/Dong-JinKim/DenseRelationalCaptioning}}.
\end{abstract}

\begin{IEEEkeywords}
Dense captioning, image captioning, visual relationship, relational analysis, scene graph.
\end{IEEEkeywords}}

\maketitle

\IEEEdisplaynontitleabstractindextext

% \fi
\IEEEpeerreviewmaketitle

%%%%%%%%%----1----%%%%%%%%%%%%%%%%%%%%%%%%%%%%%%%%%%%%%%%%%%%%%%%%%%%%%%%%%%%%%%%%%%%%%%%%%%%%%
\IEEEraisesectionheading{\section{Introduction}\label{sec:introduction}}
The human visual system has the capability to effectively and instantly collect the holistic understanding of contextual associations among objects in a scene~\cite{land2002organization,oliva2007role} by densely and adaptively skimming the visual scene through the eyes, \ie, the saccadic eye movement.
Such rich information instantly extracted from the scene allows humans to understand even subtle relationships among objects.
Motivated by {such human ability}, in this work, we present a new concept of scene understanding, called \emph{dense relational captioning} that provides dense and 
relational captions.

% Detecting objects~\cite{ren2015faster} and describing them~\cite{johnson2016densecap} are tasks that closely resemble the scene understanding of humans, which is one of the primary goals of computer vision.
% To solve this challenge, not only must the computational models perfectly detect objects, but also have the capability to capture and express their relationships in an interpretable way.
% %Each of these tasks has been regarded as a very challenging problem on its own.
% Recently, object-centric visual understanding, detection~\cite{girshick2014rich,ren2015faster} and captioning~\cite{johnson2016densecap,vinyals2015show}, has advanced significantly by deep neural networks.

%\ch{However, their capabilities are still far from the remarkable abilities of human visual system, which machine learning algorithms aim to eventually mimic.}
% \ch{The human visual system effectively collects a rich amount of contextual associations,\footnote{This is done not only by the brain, but also by the saccadic eye movements. 
%Saccadic movements provide a sampling mechanism for the brain to
%densely and adaptively skim context through visual perception~\cite{land2002organization}.
%} whereby holistic understanding of a scene in all aspects can be efficiently done~\cite{oliva2007role}.
%Such immediately extracted rich and dense information allows humans to have the superior capability for object-centric visual understanding, which has not been matched by any computer vision algorithm.}

{Rich} representation of an image often leads to performance improvements of computer vision algorithms; \eg, contexts surrounding objects of a scene~\cite{mottaghi2014role,oliva2007role}. 
To achieve richer object-centric understanding, Johnson \etal~\cite{johnson2016densecap} proposed the DenseCap {framework} that generates captions for each of densely sampled local image regions.
These regional descriptions facilitate both rich and dense semantic understanding of a scene in the form of interpretable language.
{In contrast}, the information that we want to acquire includes not only {that} of the objects itself but also the \emph{interaction} {among} other surrounding objects or the environment.

\begin{figure}[t]
	\vspace{-2mm}
	\centering
		\includegraphics[width=1.0\linewidth,keepaspectratio]{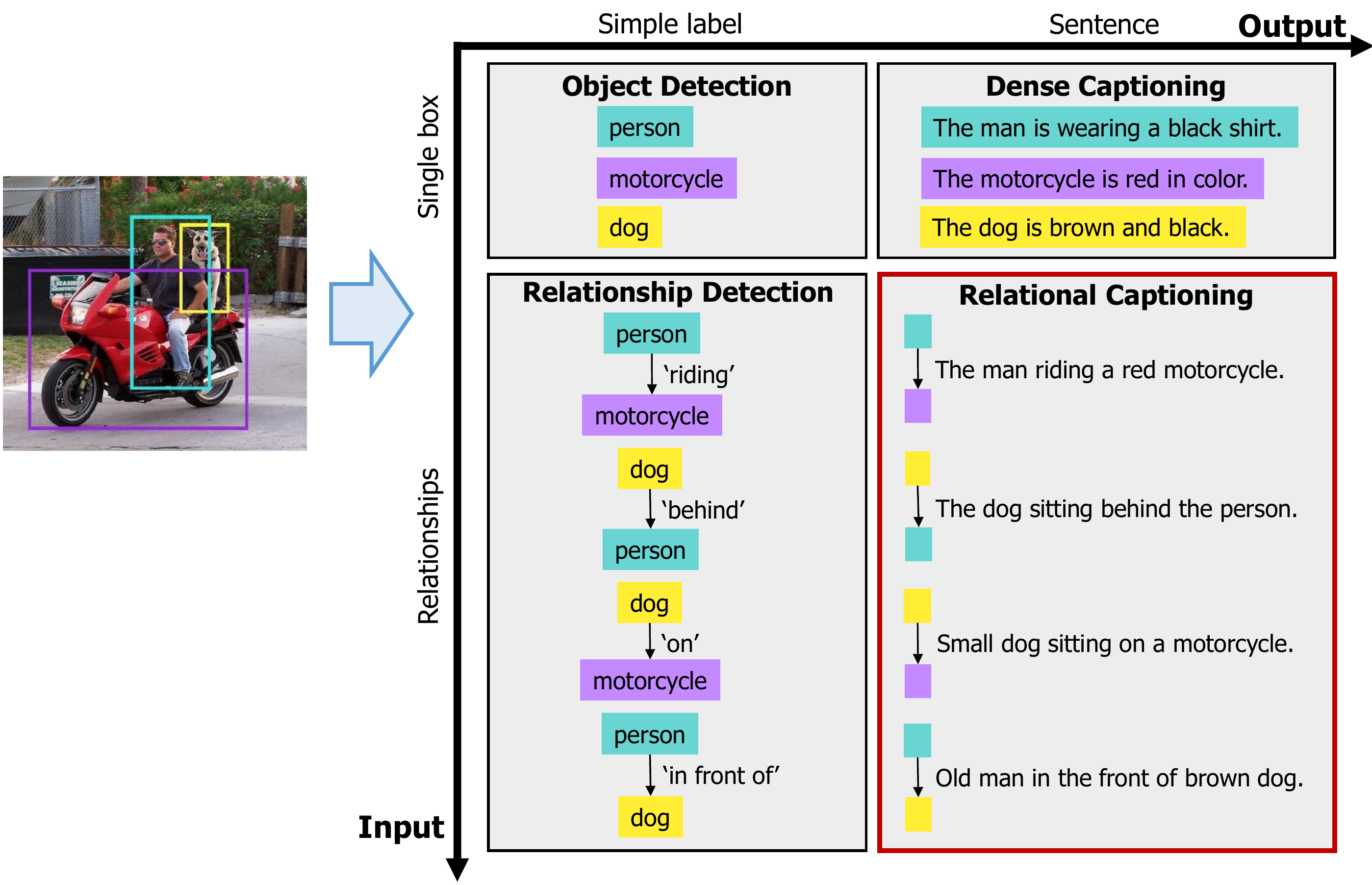}
	\vspace{-8mm}
	\caption{{Difference of our proposed relational captioning from existing image understanding frameworks.}
% 	Overall description of the proposed relational captioning framework. 
	Compared to traditional frameworks, our work is advantageous in both interaction understanding and high-level expressive interpretation.\vspace{-4mm}}
	\label{fig:teaser}
\end{figure}

As {an alternative way} of representing an image, we focus on dense \emph{relationships} between objects.
% , as objects tend to co-vary with other objects and particular environments in the real world.
In the context of human cognition, there has been a general consensus that objects and particular environments near the target object affect search and recognition efficiency.
Understanding the relationships between objects clearly reveal object interactions and object-attribute combinations~\cite{Johnson_2017_ICCV,kim2018disjoint,lu2016visual}.
%Also, relationship understanding leads to accurate retrieval and semantic understanding of the visual world~\cite{lu2016visual}.

In another view, interestingly, we observe that the human annotations on various computer vision datasets predominantly have relational forms.
%bias of human perception in computer vision datasets that annotated by human. 
In the Visual Genome~\cite{krishna2017visual} and MS COCO~\cite{lin2014microsoft} caption datasets, 
%most of the caption labels take the format of \texttt{subj-pred-obj} more so than \texttt{subj-pred}.
most of the labels take the format of subject-predicate-object more so than subject-predicate.
%\djkim{We checked the statistics of MS COCO caption dataset by using the NLTK toolkit~\cite{loper2002nltk} (POS tagging), and 88.25\% out of the generic captions has the SVO form.}
Moreover, the UCF101~\cite{soomro2012ucf101} action recognition dataset contains 85 actions out of 101 (84.2\%) that are described in terms of human interactions with other objects or surroundings. 
These aspects tell us that understanding interaction and relationships between objects facilitate a major component in object-centric visual event understanding.

In this regard, we introduce a novel captioning framework \emph{relational captioning} that can provide diverse and dense representations from a visual scene.
In this task, we first exploit the relational context between two objects as a representation unit.
This allows generating a combinatorial number of localized regional information.
Secondly, we make use of captioning and its ability to express significantly richer concepts \emph{beyond the limited label space} of object classes used in object detection tasks.
Due to these aspects, our relational captioning expands the regime further along the label space both in terms of density and complexity, and provides richer representation for an image.

% Since the generated caption describes the relationship between two objects, it should follow the sequence of subject-predicate-object.
% We leverage  to guide the correct sequence of words in a caption, so that each word is dealt with differently according to its POS category.
% we use 3 Long Short-Term Memory (LSTM)s~\cite{hochreiter1997long} each corresponding to the subject, object, and predicate, where the temporal dependencies of these LSTMs are automatically learned.
% We instill the POS prior into the deep network, called a multi-task triple-stream network (MTTSNet). 
% This model learns a multi-task of POS classification and captioning given a relation pair.

Our main contributions are summarized as follows.
(1) We introduce \emph{relational captioning}, a new {captioning} task that 
generates captions with respect to relational information between objects in an image.
(2) In order to efficiently train the relational caption information, we propose the \emph{multi-task triple-stream network} (MTTSNet) that consists of three recurrent units trained via multi-task learning with the part-of-speech prediction.
(3) %We show the effectiveness of the proposed model by evaluating on our relational captioning dataset.
%\djkim{In order to construct a relational captioning data with more elaborate sentences, we  utilize the Visual Genome (VG)~\cite{krishna2017visual} dataset and automatically augment the labels by using the attribute recognition dataset.}
We show that our proposed method is able to generate denser and more diverse captions by evaluating on our relational captioning dataset augmented from the Visual Genome (VG)~\cite{krishna2017visual} dataset as well as other relevant tasks and datasets.
(4) We demonstrate several use cases of our framework, including ``caption graphs'' which contain richer  information than conventional scene graphs.

This work is the extension of our previous conference version~\cite{kim2019dense}. We extend it in several aspects: 
We extend our architecture by adding a relational embedding module (REM) motivated by the non-local networks~\cite{wang2018non} to explicitly augment semantic meanings of surrounding objects. 
Also, we show that the REM further enhances MTTSNet in all the application scenarios we demonstrate.
In addition, we expand our experimental results and analyses to show multiple aspects of our proposed method's algorithmic behavior.

%%%%%%%%%----2----%%%%%%%%%%%%%%%%%%%%%%%%%%%%%%%%%%%%%%%%%%%%%%%%%%%%%%%%%%%%%%%%%%%%%%%%%%%%%
\section{Related Work}
Our work {mainly} relates to two topics: image captioning and relationship detection.
In this section, {we review related work on these categorized topics.}

\noindent{\textbf{Image captioning.}}
%By virtue of deep learning and the use of recurrent neural network (\eg LSTM~\cite{hochreiter1997long}) based decoders, image captioning~\cite{ordonez2011im2text} techniques have been extensively explored~\cite{vinyals2015show,donahue2015long,karpathy2015deep,lu2016knowing,mao2014deep,pu2016variational,wu2016value,xu2015show,yang2016review,you2016image}.
By virtue of deep learning and the use of recurrent neural network (\eg, LSTM~\cite{hochreiter1997long}) based decoders, image captioning~\cite{ordonez2011im2text} techniques have been extensively explored~\cite{anderson2018bottom,donahue2015long,jiang2018recurrent,karpathy2015deep,lu2016knowing,rennie2017self,vinyals2015show,xu2015show,yao2018exploring,you2016image}.
%It also provides a single image representation in a descriptive and interpretable way.
%Recent approaches are mostly attempting to obtain information from the salient region more efficiently with various forms of attention mechanisms. They cast attention over different image regions when predicting each word. 
One of the research issues in captioning is the generation of diverse and informative captions. Thus, learning to generate diverse captions has been extensively studied recently~\cite{chen2018groupcap,dai2017towards,dai2017contrastive,krause2016paragraphs,shetty2017speaking,venugopalan2017captioning,wang2017diverse,kim2019image}.
As one of the solutions, the dense captioning (DenseCap) task~\cite{johnson2016densecap} was proposed which uses diverse region proposals to generate localized descriptions, extending the conventional holistic image captioning to diverse captioning that can describe local contexts. 
%This extends the conventional holistic image captioning for a single image to diverse captioning that can describe local contexts as well.\djkim{(maybe we can merge image captioning with dense captioning?)}
{Since DenseCap generates each caption per bounding box by only relying on an internal region of the bounding box,} Yang~\etal\cite{Yang_2017_CVPR} improves the DenseCap model by incorporating a global image feature as a context cue as well as a region feature of the desired objects with late fusion.
Motivated by this, 
%Moreover, 
in order to learn dependencies of subject, object and union representations, we incorporate a triple-stream LSTM for our captioning module and further enhance the relational embedding by a non-local layer~\cite{wang2018non}.
Our triple-stream LSTM has analogies with
% is similar to
the neural module networks which have been 
% widely
used in various language-related tasks such as 
visual question answering~\cite{andreas2016neural,hu2017learning,hu2018explainable}, visual dialog~\cite{kottur2018visual}, visual grounding~\cite{liu2019learning}, captioning~\cite{tian2020image,tan2020learning,yang2019learning}, and symbolic reasoning~\cite{gupta2020neural}.
Our triple-stream LSTM can be seen as a simplified
% simple
version of a neural module network with subject, predicate, and object modules specifically designed
% specified
for our relational captioning task.
{Moreover, our \emph{relational captioning} is able to generate even more diverse caption proposals than dense captioning by considering \emph{relations} between objects.}

\noindent{\textbf{Visual relationship detection and scene graph generation.}}
Understanding visual relationships between objects have been an important concept in various tasks.
Conventional visual relationship detection (VRD) typically deals with predicting the subject-predicate-object (in short, \texttt{subj-pred-obj}).
%Its fundamental benefit is that it can make diverse relational combinations, rich contexts~\cite{lu2016visual,sadeghi2011recognition} and relational complementary information~\cite{oliva2007role}.
%%Before the formalization of the VRD task~\cite{lu2016visual}, visual relationship had been used to improve performance on various tasks~\cite{sadeghi2011recognition,choi2013understanding,desai2012detecting,gupta2008beyond,johnson2015image,yatskar2016situation}.
%As a representative work, Sadeghi~\etal\cite{sadeghi2011recognition} introduces the \emph{Visual Phrases} concept to effectively learn context-aware object recognition.
%Following works include Desai~\etal\cite{desai2012detecting} which uses phrases to facilitate action and object detection.
%However, Visual Phrases dataset consists of only a small set of $13$ common relationships each requiring enough training examples for every combination of relationship triplets.
A pioneering work by Lu~\etal\cite{lu2016visual} formalized the VRD task and provides a dataset, while addressing the subject (or object) and predicate classification models separately.
%The separate models allow the flexibility of representation and data efficiency for training than Visual Phrases. 
Their VRD dataset has also led to extensive studies on visual relationship understanding~\cite{dai2017detecting,li2017vip,plummer2016phrase,yang2018shuffle,yin2018zoom,yu2017visual,zhang2017visual,zhang2017relationship,zhuang2017towards}.
%, which typically favor the compactness of phrase detection, \ie, the task is basically boiled down to recognize appearances (\texttt{subj/obj}) and \texttt{pred} separately.
{On the other hand, similar to the VRD task, scene graph generation has started to be explored~\cite{li2018factorizable,gu2019scene,li2017scene,woo2018linknet,qi2019attentive,wang2019exploring,xu2017scene,yang2018graph,zellers2018neural}, where the task is to generate a structured graph that expresses the context relationships of a scene and provides a compact and interpretable representation of scenes.}
Moreover, human-object interaction detection task has also started to be explored recently~\cite{chao2018learning,gao2020drg,gkioxari2018detecting,kim2020detecting,kim2021acp++,li2019transferable}.

%The VRD dataset provides $100$ object classes and $70$ predicates.
%While the combinations of them are far larger than Visual Phrases task, but it is still limited to deal with exponentially large space of objects, attributes, actions, and interactions in real world. 
Although the VRD dataset is larger ($100$ object classes and $70$ predicates) than Visual Phrases~\cite{sadeghi2011recognition} dataset, it is still inadequate to handle real world scale.
%This requires compositional property~\cite{Johnson_2017_ICCV}.
The Visual Genome (VG) dataset~\cite{krishna2017visual} for relationship detection consists of $31,000$ predicate types and $64,000$ object types, which provides the combinatorial relationship triplets that are too diverse for the VRD models to comply with.
% giving the number of possible combinations of relationship triplets too diverse for the state-of-the-art VRD based models.
% {giving the combinatorial relationship triplets to the state-of-the-art VRD based models, of which number is too diverse for the VRD models to comply with.}
% {This is because the labels consist of the various combinations of words (\eg, ``little boy,'' ``small boy,'' \emph{etc.})}
{This is because, in the VRD task, each object label should be assigned to each of the various adjective and noun combinations, \eg, respective different labels for ``little boy'' and ``small boy.''}
%For example, in Figure~\ref{fig:boys}, the word ``boy'' is represented in various ways, \eg ``little boy'', ``small boy'', ``boy''.
As a result, only the simplified version of VG relationship dataset has been studied~\cite{dai2017detecting,li2017vip}. 
In contrast, our method is able to represent 
% comprehensive
extensive natural language of relations by tokenizing the whole relational expressions into words and learning from them directly.
%Given an image, our work generates combinations of object proposals followed by captioning each pair of them in the natural language form that can flexibly represent description compositionally.

While the recent state-of-the-art VRD~\cite{li2017vip,lu2016visual,plummer2016phrase,yu2017visual,yin2018zoom} {or scene graph generation~\cite{gu2019scene,li2017scene,woo2018linknet,xu2017scene,zellers2018neural}} methods 
% mostly
attempted to
use language priors to detect relationships, we directly learn the relationship 
in a descriptive language 
% model.
form.
{In addition, the expressions of the scene graph generation or the VRD tasks are restricted to \texttt{subj-pred-obj} triplets, whereas our proposed relational captioning task can provide additional information such as attributes or noun modifiers by adopting free-form natural language expressions.
Thereby, we present an extended scene graph representation, called \emph{caption graph}.
}
%They use it as a prior to transfer the knowledge of co-occurrence among subject/object and predicate.
%Contrastively, we directly learn the relationship as a descriptive language model. 

In summary, dense captioning facilitates a natural language interpretation of regions in an image, while VRD can predict relational information between objects within a restricted set. 
Our work combines both axes, resulting in much denser and more diverse captions than DenseCap.
That is, given $B$ number of region proposals in an image, we can obtain $B(B{-}1)$ number of relational captions, whereas DenseCap returns only $B$ number of captions.
This property can be favorable for subsequent algorithms in other downstream tasks.

%%%%%%%%%----3----%%%%%%%%%%%%%%%%%%%%%%%%%%%%%%%%%%%%%%%%%%%%%%%%%%%%%%%%%%%%%%%%%%%%%%%%%%%%%
\section{Multi-task Triple-Stream Networks}
\begin{figure*}[t]
\vspace{-2mm}\begin{center}		\includegraphics[width=1.0\linewidth,keepaspectratio]{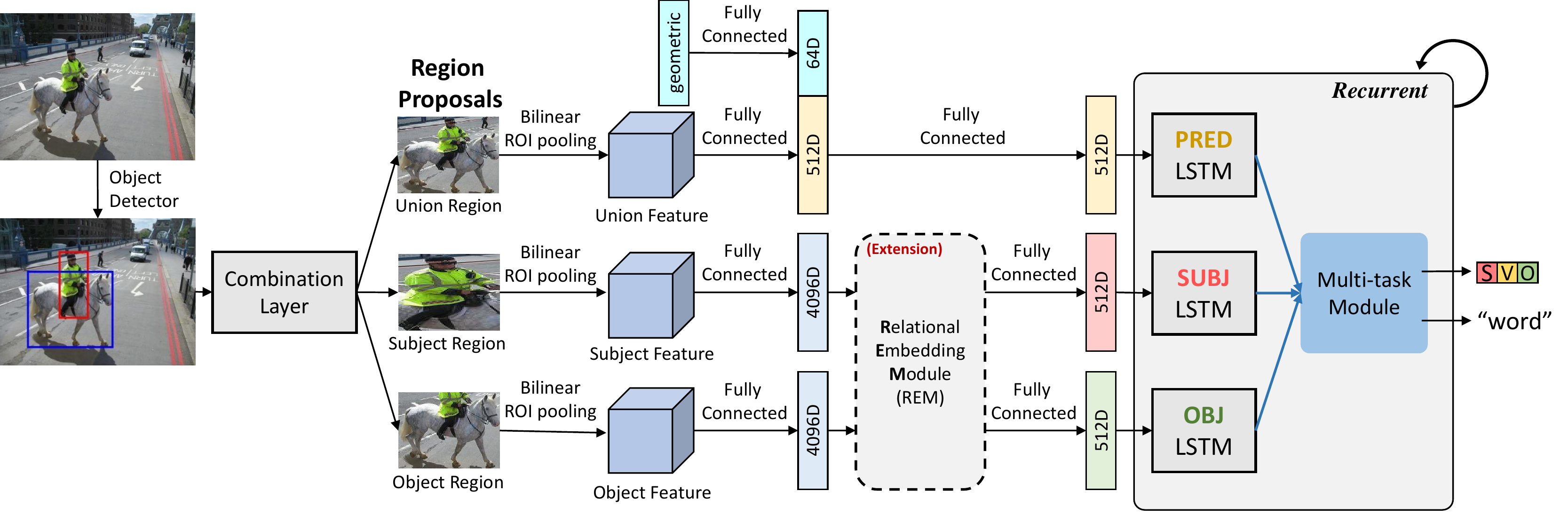}
	\end{center}
	\vspace{-6mm}
	\caption{Overall architecture of the proposed multi-task triple-stream networks (MTTSNet). 
    %The combination and region proposal layers generate combinatorial triplets for \emph{subject}, \emph{object}, and \emph{union} regions.
    Three region features (Union, Subject, Object) come from the same shared branch (Region Proposal Network), and for \emph{subject} and \emph{object} features, the first intermediate FC layer share the weights. % (depicted in the same color).
    %, while the second FC layer does not.
    %These are fed to respective LSTMs and combined through a fusion operation to jointly generate \djkim{words and it's POS for  captions}. 
    %The detail structure of the recurrent module is depicted in \Fref{fig:triple-stream}.
    %\djkim{(added relational embedding module.Intermediate feat dimension for sub/obj is changed 512$\xrightarrow{}$4096)}
    Relational Embedding Module (REM) is introduced as an extension, which takes into account early dependency between \emph{subject} and \emph{object}.
    \vspace{-4mm}
    }
	\label{fig:architecture}
\end{figure*}

Our relational captioning {generates captions} as follows. 
Given an input image, a bounding box detector generates various object proposals, {followed by a captioning module that} predicts combinatorial captions describing each pair of objects {along with} POS labels.
{This pipeline is illustrated in} \Fref{fig:architecture},
% shows the overall framework of the proposed relational captioning model, 
which is composed of a localization module based on the region proposal network (RPN)~\cite{ren2015faster}, and a triple-stream RNN (LSTM~\cite{hochreiter1997long}) module for captioning.
In addition, we introduce the relational embeddding module (REM) as an extension, to encourage explicit encoding of relational information.
Our network supports end-to-end training {within} a single optimization step that allows joint localization, combination, and description with natural language.

{Specifically,} given an image, {the} RPN generates object proposals.
Then, the combination layer takes a pair {of proposals and assigns them to the \emph{subject} and \emph{object} regions} at a time.
Also, to take the surrounding context information into account, we utilize the \emph{union} region of the \emph{subject} and \emph{object} regions as side information.
% , in a {similar way} to using the global image region as side information as done by Yang~\etal\cite{Yang_2017_CVPR}.
{These triplet features from the \emph{subject}, \emph{object}, and \emph{union} regions} are fed to the triple-stream LSTMs, where each stream takes its own purpose, \ie, \emph{subject}, \emph{object}, and \emph{union}.
Given {these triplet features}, the triple-stream LSTMs collaboratively generate a caption and POS classes of each word.
We describe details of these processes in the following sub-sections.

%-----------------------3.1--------------------------------------------------
\subsection{Region Proposal Networks}
Our network uses fully convolutional layers of VGG-16~\cite{simonyan2014very}\footnote{One can improve the performance of our relational method by replacing the backbone network with a deeper one, \eg, ResNet~\cite{he2016deep}, which we show later in the experiment section.
% In this manuscript, we do not explore this direction in that our contribution is on the introduction of relational caption representation and the relational system for it.
} up to the final pooling layer (\ie\texttt{pool5}) for extracting the spatial features via the bilinear region-of-interest (ROI) pooling~\cite{johnson2016densecap}.
The object proposals are generated by RPN~\cite{ren2015faster}.
It takes the feature tensor from the \texttt{pool5} layer, and proposes $B$ {number of} regions of interest after non-maximum suppression (NMS). 
Each proposed region {comes with} its confidence score, region feature of shape $512{\times}7{\times}7$, and coordinates $b {=} (x,y,w,h)$ of the bounding box with center $(x,y)$, width $w$ and height $h$.

Relational proposals are generated by building 
% To generate relational proposals, we build 
pairwise combinations of $B$ {number of} region proposals, where in turn we get $B(B{-}1)$ possible region pair combinations.
We call this layer {as} \emph{combination layer}.
A distinctive point of our model with the previous dense captioning 
% works
methods~\cite{johnson2016densecap,Yang_2017_CVPR} is that, while the methods regard each region proposal as an independent target to describe and produce $B$ number of captions, we consider their pairwise {$B(B{-}1)$ number of} combinations, which are much denser and explicitly expressible in terms of relationships. 
Also, we can asymmetrically use each entry of a pair by assigning the roles of the regions, \ie, (\emph{subject}, \emph{object}) or \emph{vice versa}.

{We vectorize the region features, and then apply two fully-connected (FC) layers to map them into $D$-dimensional features, where the intermediate dimensions are $D_u{=}512$ for the union region and $D_o{=}4096$ for \emph{subject} and \emph{object} regions.}
{Only the first intermediate FC layer for \emph{subject} and \emph{object} features shares their weights.}
We use the rectified linear (ReLU) units~\cite{nair2010rectified} and Dropout~\cite{srivastava2014dropout} for the FC layers.
% \djkim{For union region feature, the intermediate feature dimension is $D_u{=}512$, and for subject and object regions, the dimension is $D_o{=}4096$. (changed feat dimension for journal version.)}
{The \emph{subject} and \emph{object} region features are optionally fed to the Relational Embedding Module (REM)
which outputs refined features with the same size $D{=}512$.}
The details of the REM 
% is illustrated in \Fref{fig:extension} and will be
is described in \Sref{sec:relcapnet}.
{In short, the aforementioned process encodes region features into $D$-dimensional features, which is called \emph{region codes}.}
%Once the region codes are extracted, they are reused for the following processes.

Furthermore, {we leverage an additional region, the \emph{union} region $b_u$ of (\emph{subject}, \emph{object}) motivated by Yang~\etal\cite{Yang_2017_CVPR}.}
Yang~\etal demonstrate that the global context of an image as a side-information can improve the captioning performance.
{Compared to the global context of Yang~\etal, our union region has more localized information incorporating both subject and object.}
%We define the \emph{union} region as the smallest bounding box that covers both \emph{subject} and \emph{object} regions same as the current relationship detection works~\cite{dai2017detecting,li2017vip,lu2016visual,yu2017visual,zhang2017relationship}.
In addition, to provide relative spatial information, we append geometric features for the \emph{subject} and \emph{object} box pair, \ie, $(b_s, b_o)$, to the \emph{union} feature. 
Given two bounding boxes $b_s{=}(x_s,y_s,w_s,h_s)$ and $b_o{=}(x_o,y_o,w_o,h_o)$, we use the following  geometric feature $\mathbf{r}$ similar to {that of Peyre et al.} \cite{peyre2017weakly} as\vspace{-1mm}
\begin{equation}
        \mathbf{r} = \left[ \tfrac{x_o - x_s}{\sqrt{w_s h_s}} ,  \tfrac{y_o - y_s}{\sqrt{w_s h_s}}, \sqrt{\tfrac{w_o h_o}{w_s h_s}} , 
		\tfrac{w_s}{h_s}, \tfrac{w_o}{h_o} , 
		\tfrac{b_s \bigcap b_o}{b_s \bigcup b_o} \right] \in \R^6,\vspace{-1mm}
		\label{eq:geometric}
\end{equation}
{where $b_s \bigcap b_o$ and $b_s \bigcup b_o$ denotes the intersection and union areas of the two boxes respectively.}
The geometric feature {$\mathbf{r}$ is encoded into a 64-dimensional geometric vector by passing through an additional FC layer.} 
{By concatenating the 64-dimensional geometric vector with the \emph{union} feature, the shape of this feature is $D+64$.}
Then, the dimension of the \emph{union} region code is reduced by the following FC layer.
This stream of the aforementioned operations is illustrated in \Fref{fig:architecture}.
The three features extracted from the \emph{subject}, \emph{object}, and \emph{union} regions are fed to each LSTM described in the following sections.

\begin{figure*}[t]
	\vspace{-2mm}
	\centering
	\includegraphics[width=1.0\linewidth,keepaspectratio]{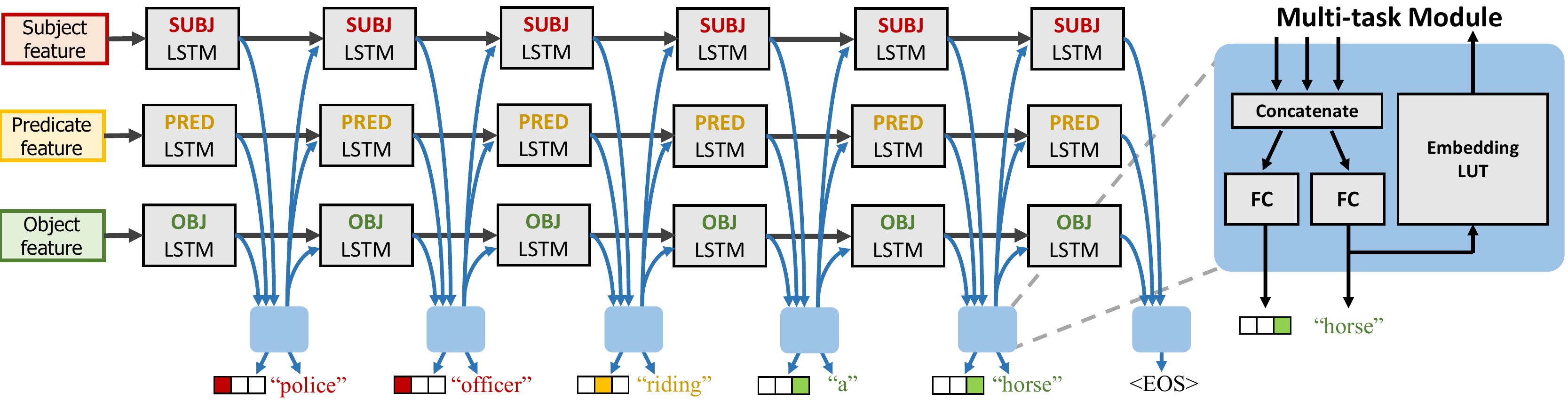}
	\caption{An illustration of the unrolled triple-stream LSTM. 
		Our model consists of two major parts: triple-stream LSTM and a multi-task module.
		The multi-task module jointly predicts a caption word and its POS class (\texttt{subj-pred-obj}, illustrated as three cells colored according to the POS class), as well as the input vector for the next time step.
        \vspace{-4mm}}
	\label{fig:triple-stream}
\end{figure*}

%-----------------------3.2--------------------------------------------------
\subsection{Relational Captioning Networks}
\label{sec:relcapnet}
{Our 
%relational captioning 
network consists of multiple LSTM modules to generate captions that describe relational information.}
To this end, we design a new network that explicitly exploits relational cues.
% {In this work, the relational information refers to the context of object pair relationships.}

{In the proposed relational region proposal, a distinctive facet is its capability to provide a triplet of region codes corresponding to the \emph{subject}, \emph{object}, and \emph{union} regions, which can be also viewed as
% represents 
the POS of a sentence  (\texttt{subj}-\texttt{pred}-\texttt{obj}).}
The existence of these correspondences between each region in a triplet and POS information can lead to the following advantages: 1) {input region codes} can be {adaptively merged} depending on their POS roles and be fed to the final word prediction module, 
and 2) when predicting a word, the POS prior can effectively affect the quality of caption generation by reducing potentially spurious words.
{To leverage these benefits, we propose the \emph{multi-task triple-stream network} (MTTSNet).
For the first advantage, to derive the POS aware inference, we propose the \emph{triple-stream network} which consists of three separate LSTMs} respectively corresponding to \texttt{subj}-\texttt{pred}-\texttt{obj}.
The outputs of the LSTMs are combined via concatenation.
For the second advantage, during word prediction, we jointly infer POS classes of each word.
This POS prediction task allows the network to learn the POS prior knowledge for the word prediction.
% for a caption.

\vspace{1mm}\noindent\textbf{Triple-Stream LSTMs.}\quad
Intuitively, the region codes of the \emph{subject} and \emph{object} would be closely related to the {respective} subject and object related words in a caption, while the \emph{union} and geometric features may contribute to the predicate.
In our relational captioning framework, the LSTM modules {need to} adaptively take {into account input features to generate a caption according to the POS decoding stage.}
% For example, when predicting a word related to \texttt{pred} in a caption, the LSTM modules give a more weight to the information from \emph{union}.

As shown in \Fref{fig:architecture}, the proposed triple-stream LSTM module consists of three separate LSTMs, each of which is in charge of the \emph{subject}, \emph{object} and \emph{union} region codes, respectively.
{From RPN, a triplet of region codes are fed as input to LSTMs,
% \footnote{{In this extension, we introduce an additional intermediate module before directly feeding the region codes to LSTMs, \emph{relational embedding module}, which is annotated by a dotted line box in \Fref{fig:architecture}. This  will be discussed later.}} 
so that a sequence of words (caption) is generated.}
At each step (word), the triple-stream LSTMs generate three embeddings separately, and a single word is predicted by consolidating the three processed embeddings {by the \emph{multi-task module} (described in the next sub-section)}.
The embedding of the predicted word is distributed into all three LSTMs as inputs of the next step and is used to run the next step in a recursive manner.
Thus in each step, each entry of the triplet input is used differently, which allows more flexibility than that of a single LSTM as used in traditional captioning models~\cite{johnson2016densecap,vinyals2015show}.
%% sequence conditional utilization of different inputs
In other words, the importance of the input features changes at every recursive step according to which POS the word being generated belongs to.

\vspace{1mm}\noindent\textbf{Multi-task with POS Classification.}\quad
At each part of the triple-stream LSTMs, we obtain three intermediate output features from each LSTM.
{To predict a word, we aggregate the features from the subject, predicate and object information, via a single FC layer.}
Also, we add an additional side task, POS prediction, from the same concatenated feature.
We call this fusion layer as the \emph{multi-task} module as shown in the right enlarged view of \Fref{fig:triple-stream}.

The multi-task module can be viewed as a \emph{late fusion} approach. {An alternative would be \emph{early fusion}, which consolidates the information in an even earlier step, \ie, the fusion of the three region codes (\eg, concatenation of three codes) followed by a single LSTM model instead of the triple-stream LSTMs.}
% However, as reported by Yang~\etal\cite{Yang_2017_CVPR}, this early fusion approach shows lower performance than that of late fusion.
However, we observe that this early fusion approach has lower performance than our late fusion one, which is also consistent with the observation reported by Yang~\etal\cite{Yang_2017_CVPR}.
Thus, we take the late fusion approach and compare the performance in \Sref{sec:exp}.

% When three representations {for each POS} are to be consolidated, 
% one option can be to consolidate them in an early step, called \emph{early fusion}.
% This results in a single LSTM with the fusion of the three region codes (\eg concatenation of three codes).
% However, as reported by Yang~\etal\cite{Yang_2017_CVPR}, this early fusion approach shows lower performance than that of late fusion.
% In this regard, we adopt \emph{late fusion} for the multi-task module.
% The layer basically concatenates the representation outputs from the triple-stream LSTMs, 

The POS classification task is leveraged to 
more effectively
train the relational captioning. 
We impose the POS classification loss during training, so that the networks learn which LSTM they should {emphasize more} at a {word prediction}.
Thereby, relational captioning is encouraged to generate a sequence of words in \texttt{subj}-\texttt{pred}-\texttt{obj} order, \ie, the order of POS.
% The POS task encourages the caption generation to follow the order of POS.
{The POS tag can be easily obtained by a modern natural language processing toolkit, NLTK POS tagger~\cite{loper2002nltk}, which had been established for a long time; thus, it provides a reliable prediction.
}
{In our case, we obtain POS (pseudo) ground truth from automatic label augmentation from relationship triplet labels.}

% but due to the recurrent multi-task modules, it is able to generate more sophisticated representations.

% {Along with the feature concatenation followed by the triple-stream LSTMs}, 

% {For each word prediction, we also classify its POS class via the \emph{multi-task} module shown in} Fig.~\ref{fig:triple-stream}, {so that it encourages the caption generation to follow the order of POS.}

%\ch{(please unify upper/lower case of `Triple-Stream')}

We empirically 
%observe 
find
that this multi-task learning with POS not only helps the shared representation to be richer, but also guides the word predictions; thus, it helps to improve the captioning performance overall. 
Since each POS class prediction relies on respective representations from each LSTM, (\eg, predicate class prediction from the \emph{pred}-LSTM), the gradients 
%generated 
from the POS classification are mainly {back-propagated} through 
{the feature elements representing {a class ambiguously} within the concatenated feature}.
Even for the same word output, the gradients from the multi-task module may differ by this fact, so that  
representations across LSTMs can be learned to be further distinctive.
Also, the 
%imposed 
POS prior may make the network suppress spurious word candidates.

{Another potential way to leverage the POS priors would be 
% One way to truly leverage the POS priors would be 
to add an additional soft attention module to select among the three features instead of concatenating them. 
We compare this attention approach 
%(denoted as \texttt{\textbf{MTTSNet (att)}}) 
with our simple concatenation~\cite{cho2021dealing}, %in \Tref{table:captioning},
where the results show that the performance of the attention approach (44.94 Recall) is lower than our concatenation (45.96 Recall) while using more number of parameters. Thus, we use the simple concatenation.
% We measured the performance with the attention module (\texttt{\textbf{MTTSNet (att)}} in \Tref{table:captioning}), 
% but the performance was similar to concatenation, while using more parameters. 
% Due to this empirical analysis, we decided to use simple concatenation.
}

% One intuition is that 
% By virtue of this mechanism, the multi-task triple-stream LSTMs {may enable effective representation learning to predict} plausible words {following the POS order} for each time step.

% We hypothesize that the POS task provides distinctive information that may help learn proper representations from the triple-stream LSTMs.

\begin{figure}[t]
\vspace{-2mm}
\centering
\includegraphics[width=0.9\linewidth,keepaspectratio]{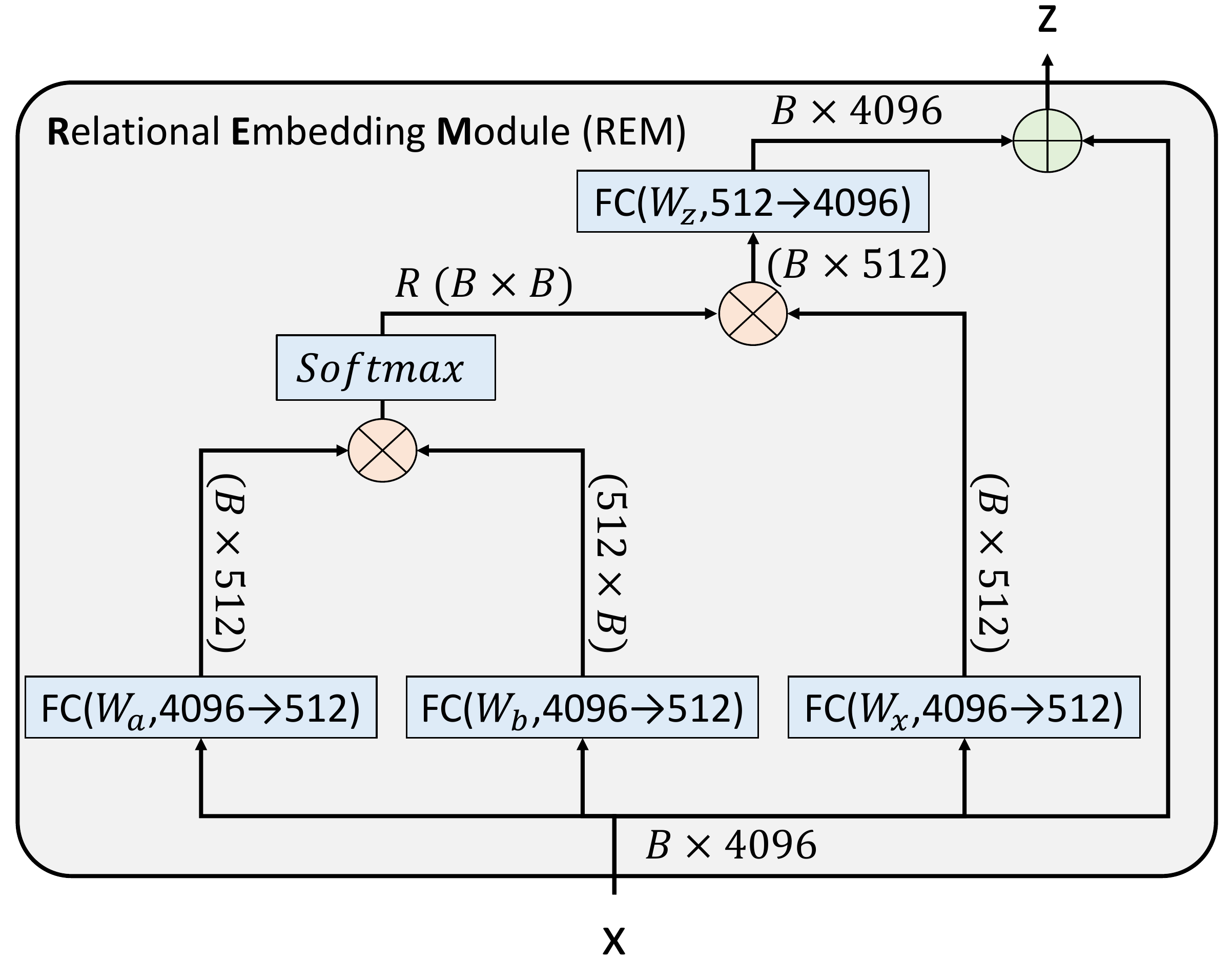}
\vspace{-2mm}
\caption{{Architecture of the relational embedding module (REM). $\bigotimes$ denotes the matrix multiplication, and $\bigoplus$ the element-wise sum. The softmax operation is applied row-wise. The blue boxes with the FC label denote FC layers.}
}
\vspace{-4mm}
\label{fig:extension}
\end{figure}

\vspace{1mm}\noindent\textbf{Relational Embedding Module.}\quad 
Since our triple-stream network only utilizes the triplet features (subject, object, and union), it alone may lack global understanding of the constituent objects in an entire image, \ie, global context.
In this extension, to strengthen the capability of holistic relational understanding across all the objects, we employ the non-local layer~\cite{hu2018relation,kim2021acp++,wang2018non},
% \djkim{or self-attention layer~\cite{hu2018relation}}, 
we called the relational embedding module (REM), where we apply the non-local layer to each object candidate.
This is different from Wang~\etal\cite{wang2018non}, where they apply it to the feature map densely while we apply it to ROI pooled features.
The REM enhances the relational information across all objects via the attention mechanism.

Specifically, let $X \in \R^{B\times D_o}$ denote a stack of $B$ number of vectorized region features extracted from the first FC layer after the bilinear ROI pooling.
% \djkim{(newly added.)}
% With the previously published version of our triple-stream network, the caption module {lacks} global {knowledge} of the {constituent} objects {in an entire image}.
% As an extension of our previously published work, we {employ global context knowledge to} further improve our triple-stream by leveraging the Relational Embedding Module suggested by Woo~\etal~\cite{woo2018linknet}.
% Let us denote the vectorized region features, that passed the first FC layer after bilinear ROI pooling as $X \in \R^{B\times D_o}$.
% {Given the} feature matrix $X$, 
Then, we compute the relational association matrix by:
\begin{equation}
        R = \mathrm{softmax}({\sigma}(X W_a){\sigma}(X W_b)^\top) \in \R^{B\times B},
		\label{eq:REM1}
\end{equation}
where {$\sigma(\cdot)$ denotes ReLU and} $W_a, W_b \in \R^{D_o\times 512}$ are learnable {weights} that map the region features $X$ to each of its own role, (\eg, subject and object) and 
the softmax operation is applied in row.
Then, the relational feature matrix is computed by:
\begin{equation}
A = R\sigma(XW_x)W_z^\top \in \R^{B\times D_o},
\label{eq:REM2}
\end{equation}
where $W_x\in \R^{D_o\times 512}$ and $W_z\in \R^{D_o\times 512}$ are again learnable {weights}.
The matrix $A$ encodes aggregated features across all the objects according to the degree of relational association by $R$, which is similar to the message passing that exchanges the information according to relationships.
%  and  are that map the region features $X$ to $\R^{B\times 512}$ {and  is the weights for mapping to $\R^{B\times D_o}$}.
{This relational feature matrix is combined with the original feature $X$ by $Z = X + A$, so that the holistic relational information is enhanced on top of $X$.
This can be viewed as augmenting richer semantic meanings, \eg, a shirt ($X$) is augmented to a shirt that someone is in or a shirt on something depending on the surroundings.
Also, it is akin to the residual connection, allowing efficient training via the residual learning mechanism~\cite{he2016deep}.
{Different from the non-local approaches \cite{wang2018non,woo2018linknet}, we introduce non-linear activations, ReLU, in Eqs.~(\ref{eq:REM1}) and (\ref{eq:REM2}),  motivated by a low-rank bilinear pooling method~\cite{kim2016hadamard}.
We empirically found this modification leads to noticeable performance improvement.}

{Furthermore, similar to Hu~\etal\cite{hu2018relation}, we may leverage the box geometric features $\mathbf{r}$ in the REM to re-scale the attention. However, in our empirical experiment, this does not help and even lowers the performance than our method that concatenates geometric features to the union feature. Thus, we use the final REM module illustrated in \Fref{fig:extension}.}
% \djkim{Also, , we tried to leverage the box geometric features $\mathbf{r}$ to re-scale the attention weight $A$. 
% The REM module is illustrated in \Fref{fig:extension}.
}

% The goal of the relational embedding module is to refine this matrix $X$ to a matrix $Z$ that has the same shape with $X$.
% {Inspired by recent works on relational reasoning~\cite{santoro2017simple,vaswani2017attention,wang2018non,woo2018linknet}, we compute object-relational embeddings $R$ that computes the response for one object region by attending to all $B$ object regions. }
%\js{In particular we follow the method of Woo et al. that considers object-level instances as the primitive elements, whereas the previous methods operate on pixels or words.}

%\noindent\textbf{Label noise handling.}\quad
%\djkim{contents about label noise handling will be added }

\noindent\textbf{Loss functions.}\quad
% Training {loss of} our relational captioning model can be mainly divided into captioning loss and detection loss. 
% Specifically, t
The proposed model is trained to minimize the following loss function:
\begin{equation}
		\mathcal{L} = \mathcal{L}_{cap} + \alpha \mathcal{L}_{POS} + \beta \mathcal{L}_{det} + \gamma \mathcal{L}_{box},
		\label{eq:loss}
\end{equation}
where $\mathcal{L}_{cap}$, $\mathcal{L}_{POS}$, $\mathcal{L}_{det}$, and $\mathcal{L}_{box}$ denote captioning loss, POS classification loss, detection loss, and bounding box regression loss, respectively.
$\alpha$, $\beta$, and $\gamma$ are the balance parameters (we set them to $0.1$ for all experiments). 
The first two terms are for captioning and the next two terms are for the region proposal.
$\mathcal{L}_{cap}$ and $\mathcal{L}_{POS}$ are cross-entropy losses applied to each word and POS {prediction} at every time step, respectively. 
For each time step, $\mathcal{L}_{POS}$ measures a 3-class cross entropy loss.
$\mathcal{L}_{det}$ is a binary logistic loss for foreground/background regions {to distinguish positive and negative object regions~\cite{girshick2015fast,johnson2016densecap}}, 
while $\mathcal{L}_{box}$ is a smoothed L1 loss ~\cite{ren2015faster}.

%%%%%%%%%----4----%%%%%%%%%%%%%%%%%%%%%%%%%%%%%%%%%%%%%%%%%%%%%%%%%%%%%%%%%%%%%%%%%%%%%%%%%%%%%
\section{Experiments}
\label{sec:exp}

In this section, we provide the experimental setups, competing methods and performance evaluation of relational captioning with both quantitative and qualitative results, so that we empirically show the benefit and potential of the proposed relational captioning task and the proposed method.
%-----------------------4.1--------------------------------------------------

\begin{table}[t]
    \centering
    \vspace{-0mm}
    \resizebox{1.0\linewidth}{!}{%
		\begin{tabular}{l  cccc } 
        \toprule
			Model & Output of RPN & Input of LSTM & LSTM & POS prediction\\
			\midrule
			\texttt{Direct Union} 	& Union region 	& U		& 	Single	& 	$\times$	\\
			\texttt{Union} 			& Object 		& U		& 	Single	& 	$\times$	\\
			\texttt{Union+Coord.} 	& Object 		& U + C		& 	Single	& 	$\times$	\\
			\texttt{Subj+Obj}  		& Object 		& S + O		& 	Single	& 	$\times$	\\
			\texttt{Subj+Obj+Coord.}& Object 		& S + O + C		& 	Single	& 	$\times$	\\
			\texttt{Subj+Obj+Union} & Object 		& S + O + U		& 	Single	& 	$\times$	\\
			\texttt{Union (w/MTL)} 	& Object 		& U		& 	Single	& 	$\bigcirc$	\\
			\texttt{Subj+Obj+Coord.(w/MTL)}& Object 		& S + O + C		& 	Single	& 	$\bigcirc$	\\
			\texttt{Subj+Obj+Union (w/MTL)} & Object 		& S + O + U		& 	Single	& 	$\bigcirc$	\\
			\texttt{Union+Union+Union (w/MTL)} & Object 		& U + U + U		& 	Triple	& 	$\bigcirc$	\\
			\texttt{TSNet} 		& Object 		& S $|$ O $|$ U	+ C	& 	Triple	& 	$\times$	\\
			\texttt{MTTSNet} 		& Object 		& S $|$ O $|$ U + C		& 	Triple	& 	$\bigcirc$	\\
            \bottomrule
		\end{tabular}
    \hspace*{0mm}}
    \vspace{0mm}
    \caption{Comparison of model configurations. 
    `$|$' and `+' indicate separation and concatenation of input respectively.
    }
	\vspace{-8mm}
	\label{table:baselineconf}	
\end{table}

\subsection{Experimental Setups}
\label{sec:exp_setup}
\noindent\textbf{Implementation details.}
We use Torch7~\cite{collobert2011torch7} to implement our model.
For the {backbone} visual feature extraction, we use VGG-16~\cite{simonyan2014very} and initialize with the weights pre-trained on ImageNet~\cite{russakovsky2015imagenet}.
We pre-train the RPN on the Visual Genome (VG) dense captioning data~\cite{krishna2017visual}. 
For sequence modeling, we set the {dimension} of 
all the LSTM hidden layers
% the hidden layers of all LSTMs
to be 512.
%Then, we train the full model in a single step optimization. 
A training batch contains an image that is resized to have a longer side of 720 pixels. 
We use Adam optimizer~\cite{ba2015adam} for training (learning rate $lr {=} 10^{-6}$, $b1{=}0.9$, $b2{=}0.999$).
For the RPN, we use 12 anchor boxes for generating the anchor positions in each cell of the feature map, and 128 boxes are sampled in each forward pass of training.
We use Titan X GPU, and it takes about four days for a model to convergence when training on our relational captioning dataset.

We use the setting for the region proposals similar to that of \cite{johnson2016densecap} for fairness.
For training, a region is positive if it has 
{at least 0.7 IoU ratio with a corresponding ground truth region,}
% an IoU ratio with a corresponding ground truth region of at least 0.7,
and 
% a region is negative if  it has IoU less than 0.3 with all ground truth regions. 
{a region is negative if its IoUs are less than 0.3 with all ground truth regions.}
For evaluation, after non-maximum suppression (NMS) based on the predicted proposal confidences, 50 confident bounding boxes are selected.  
%We use the beam-1 search to produce region descriptions, where the word with the highest probability is selected at each time step.  
{We can additionally reduce box pair predictions by discarding the pairs that produce captions with low 
%overall 
confidence scores.
Caption confidence scores can be computed by sequentially multiplying all of the generated word probabilities.}

%For evaluation, we use the similar setting to that of \cite{johnson2016densecap} for fairness.
%First, after non-maximum suppression (NMS) with IoU ratio 0.7, 50 confident {bounding} boxes are selected. %\oh{COMMENT: how top? how many?}
%We use the beam-1 search to produce region descriptions, where the word with the highest probability is selected at each time step. 
%Then, with another round of NMS with IoU ratio 0.3, the remaining regions and their descriptions are generated as final results.
%\oh{COMMENT: I don't understand the procedure in the last sentence and why we need it}

%\oh{COMMENT: please describe how you can reject the caption when the caption has a low confidence.}

\begin{table}[t]
\vspace{0mm}
    \centering
    \resizebox{1.0\linewidth}{!}{%
        %\resizebox{0.5\linewidth}{!}{%
         \begin{tabular}{l ccc}\toprule
										&	mAP (\%) 	&Img-Lv. Recall	&METEOR\\\midrule
            \texttt{Direct Union}		&		--	&	17.32	&11.02\\\midrule
            \texttt{Union}				&	0.57	&	25.61	&12.28\\
            \texttt{Union+Coord.}		&	0.56 	&	27.14 	&13.71\\
            \texttt{Subj+Obj}			&	0.51 	& 	28.53	&13.32\\
           \texttt{Subj+Obj+Coord.}		&	0.57 	&	30.53 	&14.85\\
           \texttt{Subj+Obj+Union}		& 0.59		& 	30.48	& 15.21\\
           \texttt{\textbf{TSNet (Ours)} }	&	\textbf{0.61}	& 	\textbf{32.36}		&\textbf{16.09}\\
           \midrule
           \texttt{Union (w/MTL)}		&	0.61	&	26.97 	&12.75\\
           \texttt{Subj+Obj+Coord (w/MTL)}		&	0.63 	&	31.15 	&15.31\\
           \texttt{Subj+Obj+Union (w/MTL)}		& 0.64	& 	31.63		& 16.63\\
            %\texttt{\textbf{MTTSNet (Ours)(sum)} } &{0.68}&{32.77}	&{16.19}\\
            \texttt{Union+Union+Union (w/MTL)}		& 0.58		& 	34.11	& 14.69\\
            \texttt{\textbf{MTTSNet (Ours)} } &{0.88}&{34.27}	&\textbf{18.73}\\
            %\hline
            %\hline
            %\texttt{\textbf{MTTSNet (Ours)(sum) + REM~\cite{woo2018linknet} }} &{0.73}&\textbf{42.82}	&{17.40}\\
            %\midrule
            %\texttt{Union+Union+Union (w/MTL)+ REM}		& 0.59		& 	35.99	& 15.45\\
            \texttt{\textbf{MTTSNet (Ours) + REM}~\cite{wang2018non}}&\textbf{1.12}&\textbf{45.96}	&{18.44}\\
            %\texttt{\textbf{MTTSNet (att) + REM}~\cite{wang2018non}}&{0.89}&{44.94}	&{18.11}\\
            \midrule
            \texttt{\textbf{MTTSNet (Ours) + REM (R)}}&{1.48}&{48.56}	&{19.48}\\
            \midrule
            \texttt{Neural Motifs}~\cite{zellers2018neural} &0.25 & 29.90&15.34\\
            %\texttt{Neural Motifs$\dagger$~\cite{zellers2018neural}  } & 0.32&31.18 &21.97\\
            \bottomrule
		\end{tabular}
        %}
       % \caption{}\label{table:captioning1b}
        
        }
        \vspace{-0mm}
	\caption{Ablation study for {the} relational dense captioning task on {the} relational captioning dataset. 
	{The second and third row sections (2-7 and 8-12th rows) show the comparison of the baselines with and without {the} POS classification (\texttt{w/MTL})}.
	In the last row, we show the performance of the state-of-the-art scene graph generator, \texttt{Neural Motifs}~\cite{zellers2018neural}.
	{Union+Union+Union denotes the results of using three LSTMs with only union features 
% 	for
	as LSTM inputs, (R) indicates ResNet-50~\cite{he2016deep} as a backbone network instead of VGG-16.} 
    }
    \vspace{-4mm}
	\label{table:captioning}	
\end{table}

\noindent\textbf{Relational captioning dataset.}
{Since there is no existing dataset for the relational captioning task, we construct a dataset by utilizing VG relationship dataset version 1.2~\cite{krishna2017visual} which consists of 85,200 images with 75,456/4,871/4,873 splits for train/validation/test sets respectively.
We tokenize the relational expressions {into word level tokens}, 
and we assign the POS class from the triplet association for each word. %, {which will be explained in detail}.
}

However, the VG relationship 
% datasets show 
{dataset has a}
limited diversity {of} the words used.
% Therefore, by only using relational expressions to construct data, the captions generated from a model tends to be simple (\eg, ``building-has-window'').
{Therefore, na\"ively converting such VRD dataset to a captioning dataset is not desirable, in that the captions generated from a trained model on the dataset tends to be too simple (\eg, ``building-has-window'').}
This limited data restricts the expressiveness of the model.
% {Even though our model may enable richer concepts and expressions,} if the training data does not contain such concepts and expressions, there is no way to actually \emph{see} this.
To examine the diverse expressions of our relational captioner, we construct our relational captioning dataset to have more natural sentences with richer expressions.

Through observation, {we noticed that labels in the VG relationship dataset lack \emph{attributes} describing the subject and object, which are perhaps what enriches the expressiveness of sentences the most}.
{We enrich the dataset by leveraging the {VG \emph{attribute} dataset}~\cite{krishna2017visual}}.
The specific procedure of {this attribute enrichment} is described in Appendix.
After this enrichment, we obtain 15,595 {different} vocabularies for our relational captioning dataset, which was 11,447 {different} vocabularies before this process. 

%Therefore, we utilize the \emph{attribute labels} of VG data \oh{COMMENT: is it the same dataset with VG relationship dataset version 1.2??
%The name should be consistent with the previous one.
%Otherwise, reference is needed and mention it is a separate dataset. }
%to augment existing relationship expressions.

{We train our model with this dataset, and report its result in this section.}
In {the following subsections}, we {evaluate in multiple views including}
% provide 
a holistic image captioning performance and various analysis such as comparison with scene graph generation.

%-----------------------4.2--------------------------------------------------
\subsection{Relational Dense Captioning: Ablation Study}
\label{sec:exp_relation}

\noindent\textbf{Baselines.} Since no direct {related} work for relational captioning exists, we implement several baselines by modifying the most relevant methods, which facilitate our ablation study.
All the configurations are summarized in \Tref{table:baselineconf} and described as follows.

{\setdefaultleftmargin{3mm}{}{}{}{}{}
\begin{itemize}
\item \texttt{Direct Union} has the same architecture with \texttt{DenseCap}~\cite{johnson2016densecap}, but of which RPN is trained to directly predict union regions.
{A union region is converted to a {512-dimensional} region code, followed by a single LSTM to generate a relational caption.}\vspace{0mm}

\item \texttt{Union} also resembles \texttt{DenseCap}~\cite{johnson2016densecap} and \texttt{Direct union}, but its RPN predicts individual object regions. 
The object regions are paired as (subject, object), and then {only} a union region from each pair is fed to a single LSTM for captioning.
Also, we implement two additional variants: \texttt{Union (w/MTL)} additionally predicts the POS classification task, and 
\texttt{Union+Coord.} appends the geometric feature to the region code of the union.
{Lastly, to match the number of parameters with our \texttt{MTTSNet}, we additionally introduce the \texttt{Union+Union+Union} baseline with the triple-stream architecture, which only takes the union region as input.}
% \djkim{Finally, in order to match the number of parameters with our \texttt{MTTSNet}, we additionally introduce \texttt{Union+Union+Union} baseline with the triple-stream architecture, which only takes the union region as input.}
\vspace{0mm}

\item \texttt{Subj+Obj} and \texttt{Subj+Obj+Union} models use the concatenated region code of (subject, object) and (subject, object, union) respectively and pass them through a single LSTM (an early fusion approach).
%The concatenated codes are fed to a single LSTM. 
Also, \texttt{Subj+Obj+Coord.} uses the geometric feature instead of the region code of the union.
Moreover, we {evaluate the baselines}, \texttt{Subj+Obj+\{Union,Coord\}} {again by adding} the POS classification (\ie, MTL loss).
\vspace{0mm}

\item \texttt{TSNet} denotes the proposed triple-stream LSTM model without a branch for the POS classifier.
Each stream takes the region codes of (subject, object, union{+}coord.) separately. 
%This is an extension of the dual-stream model~\cite{Yang_2017_CVPR} to fit the relational captioning task.\vspace{1mm}
\texttt{MTTSNet} (\ie, \texttt{TSNet}{+}POS) denotes the {multi-task triple-stream network with the POS classifier, and} \texttt{MTTSNet+REM}  denotes the model {combined with} the REM.

\end{itemize}
}
 
%Each model takes near three to four days to train.

\begin{table}[t]
\centering
    \resizebox{1\linewidth}{!}{%
		\begin{tabular}{l cc cc}\toprule
													&		Recall	&	METEOR		&\#Caption	&Caption/Box	\\\midrule
			Image Cap. (\texttt{Show\&Tell})~\cite{vinyals2015show} 	            &		23.55	&	8.66		&	1		&	N/A		\\
			Image Cap. (\texttt{Show\&Tell})~\cite{vinyals2015show}$^\dagger$ 	&		23.81	&	9.46		&	10		&	N/A		\\
            Image Cap. (\texttt{SCST})~\cite{rennie2017self} 	                &		24.04	&	14.00		&	1		&	N/A		\\
            Image Cap. (\texttt{SCST})~\cite{rennie2017self}$^\dagger$ 	        &		24.17	&	13.87		&	10		&	N/A		\\
            Image Cap. (\texttt{RFNet})~\cite{jiang2018recurrent} 	            &		24.91	&	17.78		&	1		&	N/A		\\
            Image Cap. (\texttt{RFNet})~\cite{jiang2018recurrent}$^\dagger$ 	    &		25.26	&	17.83		&	10		&	N/A		\\
            \midrule
			Dense Cap. (\texttt{DenseCap})~\cite{johnson2016densecap}    &		42.63	&	19.57	&		9.16	&	1		\\
			Dense Cap. (\texttt{TLSTM})~\cite{Yang_2017_CVPR}            &		43.15	&	20.48	&	9.24	&	1		\\
			\midrule
			Relational Cap. (\texttt{Union})	&		38.88	&	18.22	&	85.84	&	9.18	\\
			Relational Cap. (\textbf{\texttt{MTTSNet}})		&{46.78} &{21.87}	&{89.32}&{9.36}\\
			Relational Cap. (\textbf{\texttt{MTTSNet+REM}})		&\textbf{56.52} &\textbf{22.03}	&{80.95}&{9.24}\\
			\midrule
			Relational Cap. (\textbf{\texttt{MTTSNet+REM} (R)})		&{59.71} &{23.27}	&{85.37}&{9.26}\\
			\midrule
			Relational Cap. (\texttt{Union})$^{(GT)}$	&		41.64	&	18.90	&	\multirow{3}{*}{83.44}&\multirow{3}{*}{9.30}	\\
			Relational Cap. ({\texttt{MTTSNet}})$^{(GT)}$		&{48.50} &{21.63}	\\
			Relational Cap. ({\texttt{MTTSNet+REM}})$^{(GT)}$		&{56.62} &{22.50}	\\\bottomrule
		\end{tabular}
    }
    \vspace{0mm}
	\caption{Comparisons of the holistic level image captioning. We compare the results of the relational captioners with those of three image captioners~\cite{jiang2018recurrent,rennie2017self,vinyals2015show} and two dense captioners~\cite{johnson2016densecap,Yang_2017_CVPR}.
	{To compare with stronger baselines, we modify the image captioners 
	%to improve the diversity of results 
	by deploying a stochastic sampling.
	We annotate the modified versions with stochastic sampling with $\dagger$.
	We annotate $(GT)$ for the methods that replace RPN with ground truth bounding boxes; thus, those represent proxy upper bounds of performance.
	{(R) indicates ResNet-50~\cite{he2016deep} as a backbone network instead of VGG-16.}
% 	\djkim{(R) indicates the results when using ResNet-50~\cite{he2016deep} instead of VGG-16.}
	}
% 	\vspace{-6mm}
    \vspace{-4mm}
	}
	\label{table:recall}	
\end{table}

\begin{figure*}[t]
\vspace{-2mm}
\centering
%\resizebox{1\linewidth}{!}{%
{\includegraphics[height=0.24\linewidth,keepaspectratio]{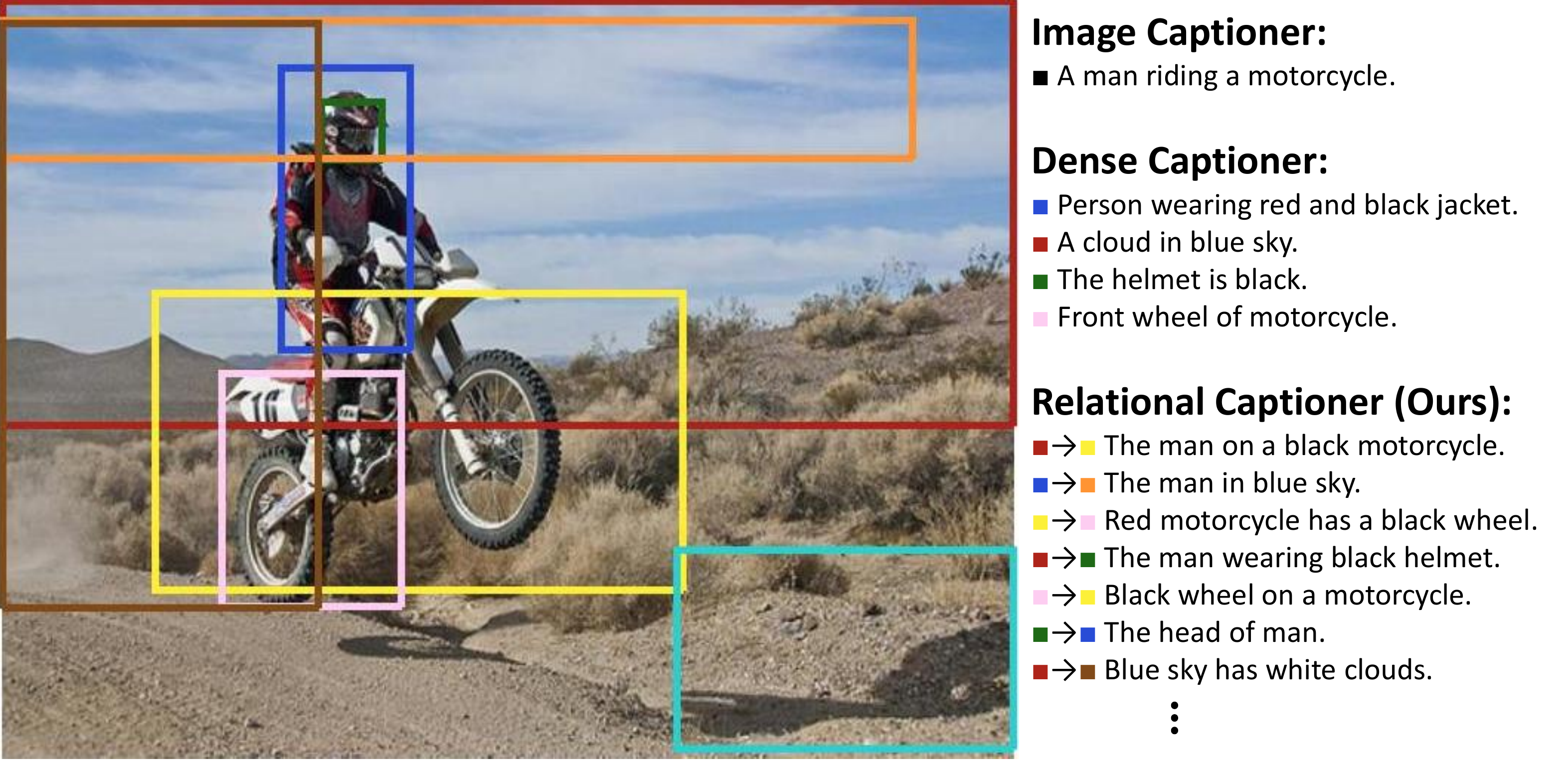}}
%{\includegraphics[height=0.5\linewidth,keepaspectratio]{Figures/qualitative2}}
{\includegraphics[height=0.24\linewidth,keepaspectratio]{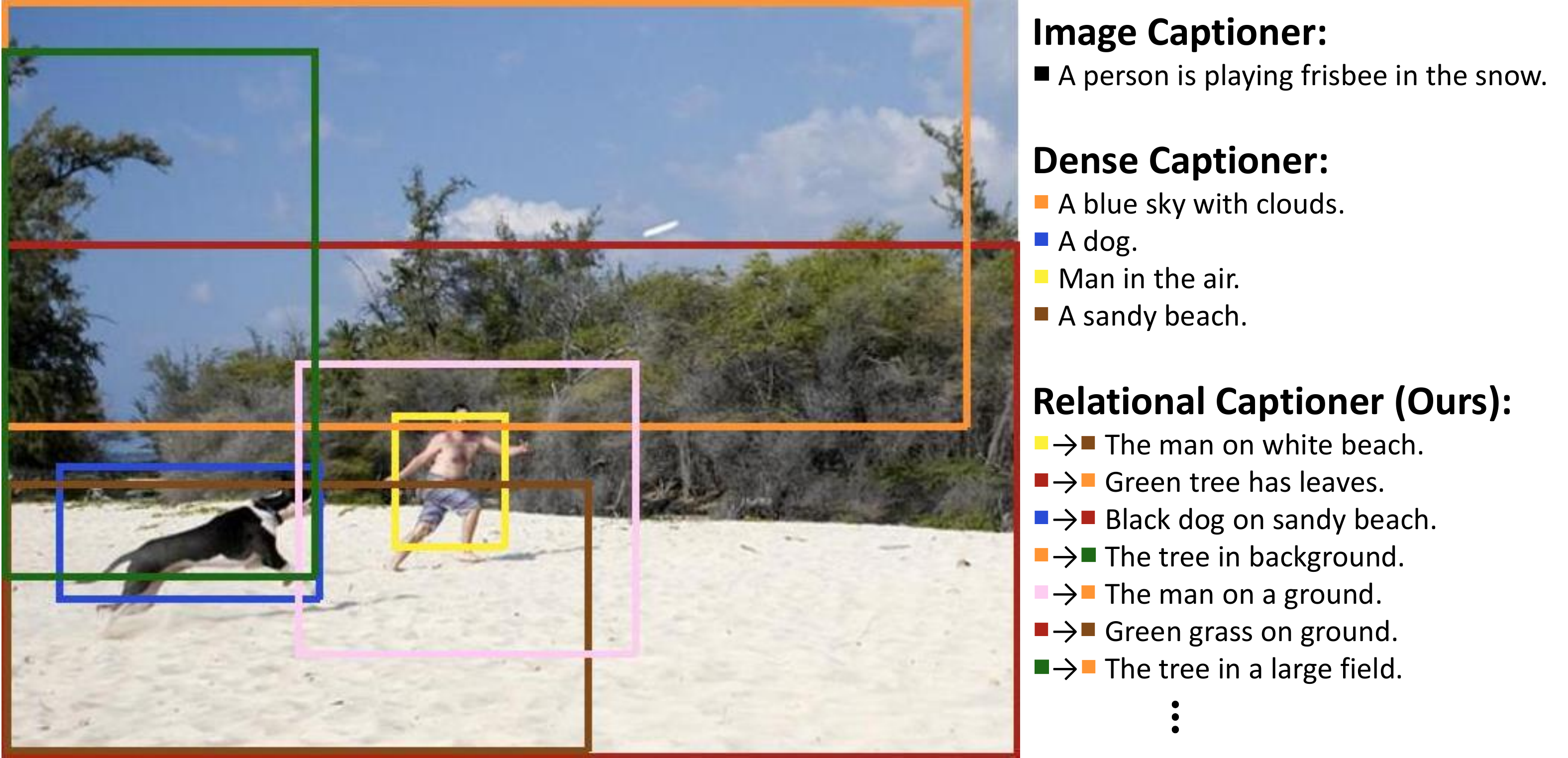}}\\
{\includegraphics[height=0.24\linewidth,keepaspectratio]{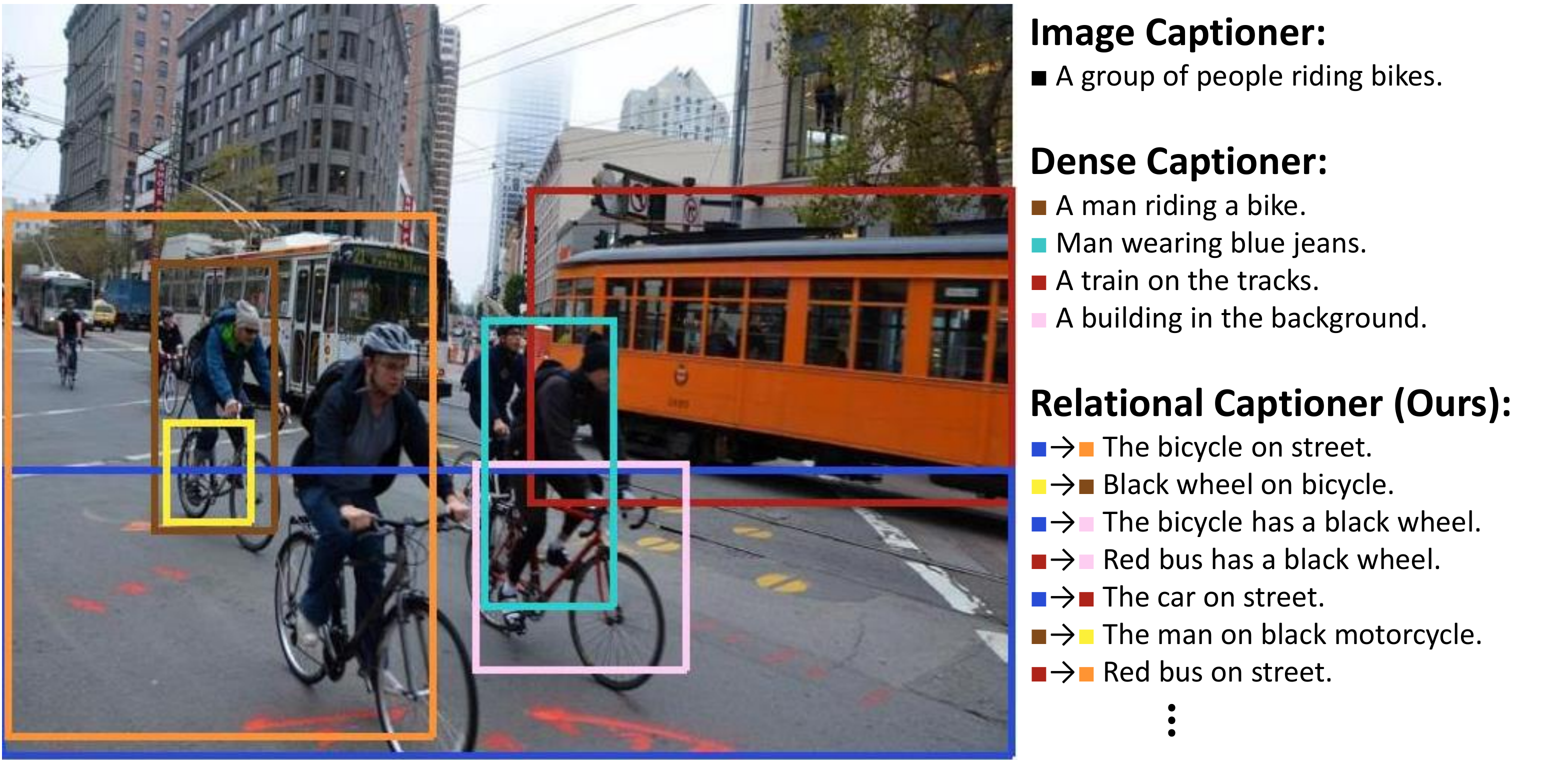}}%height=0.245 / width=0.5
{\includegraphics[height=0.24\linewidth,keepaspectratio]{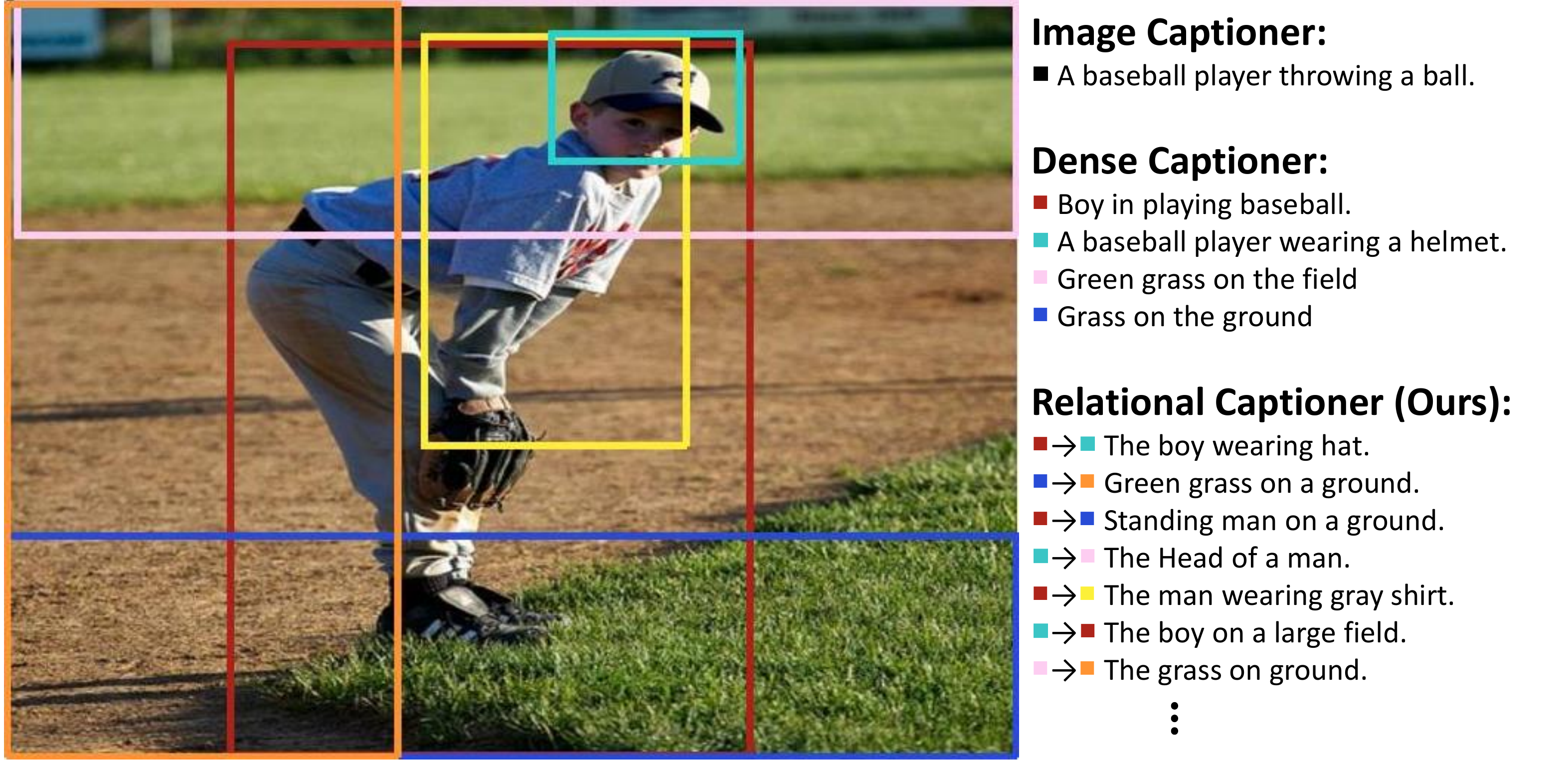}}\\
{\includegraphics[height=0.24\linewidth,keepaspectratio]{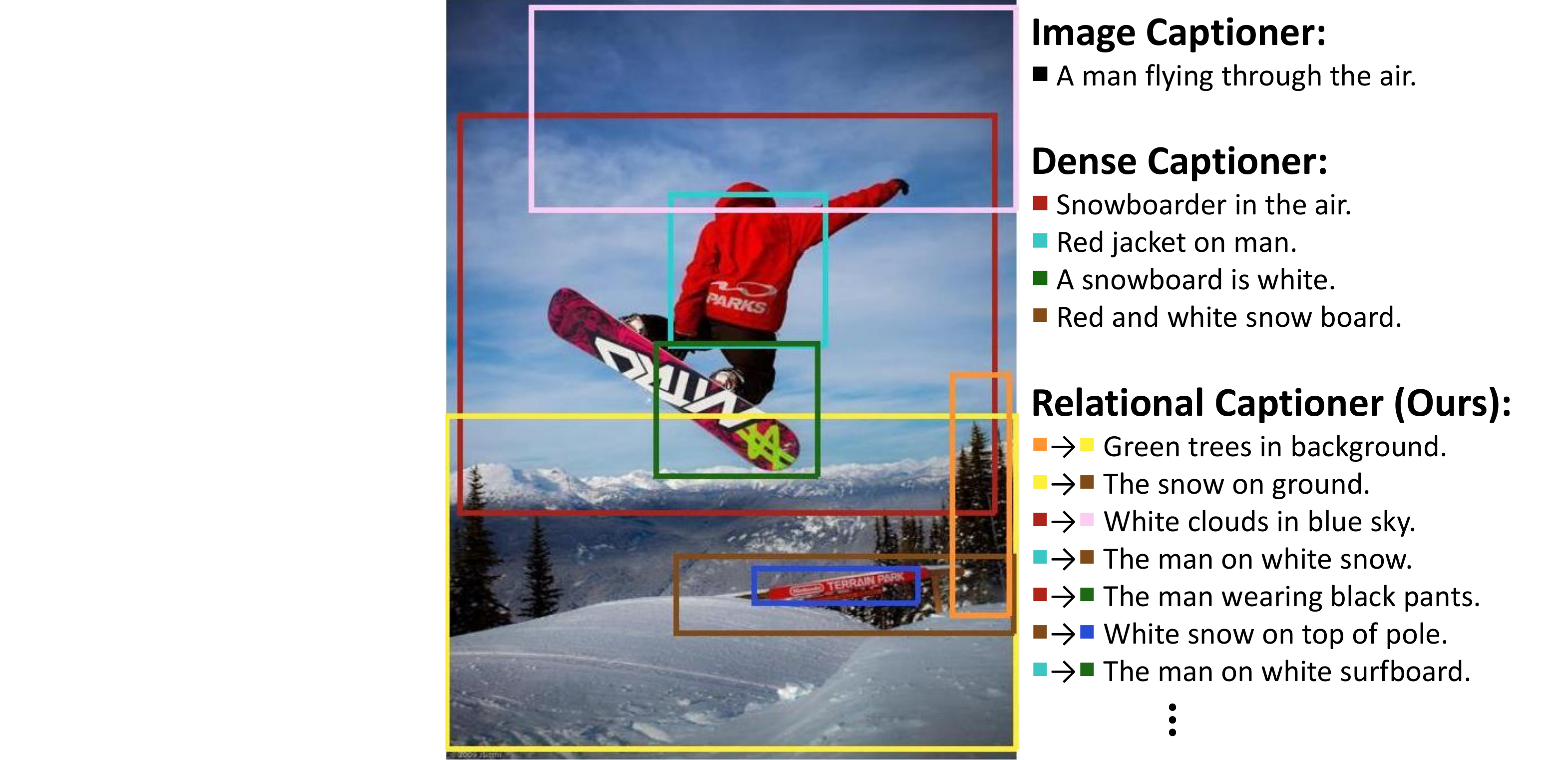}}
{\includegraphics[height=0.24\linewidth,keepaspectratio]{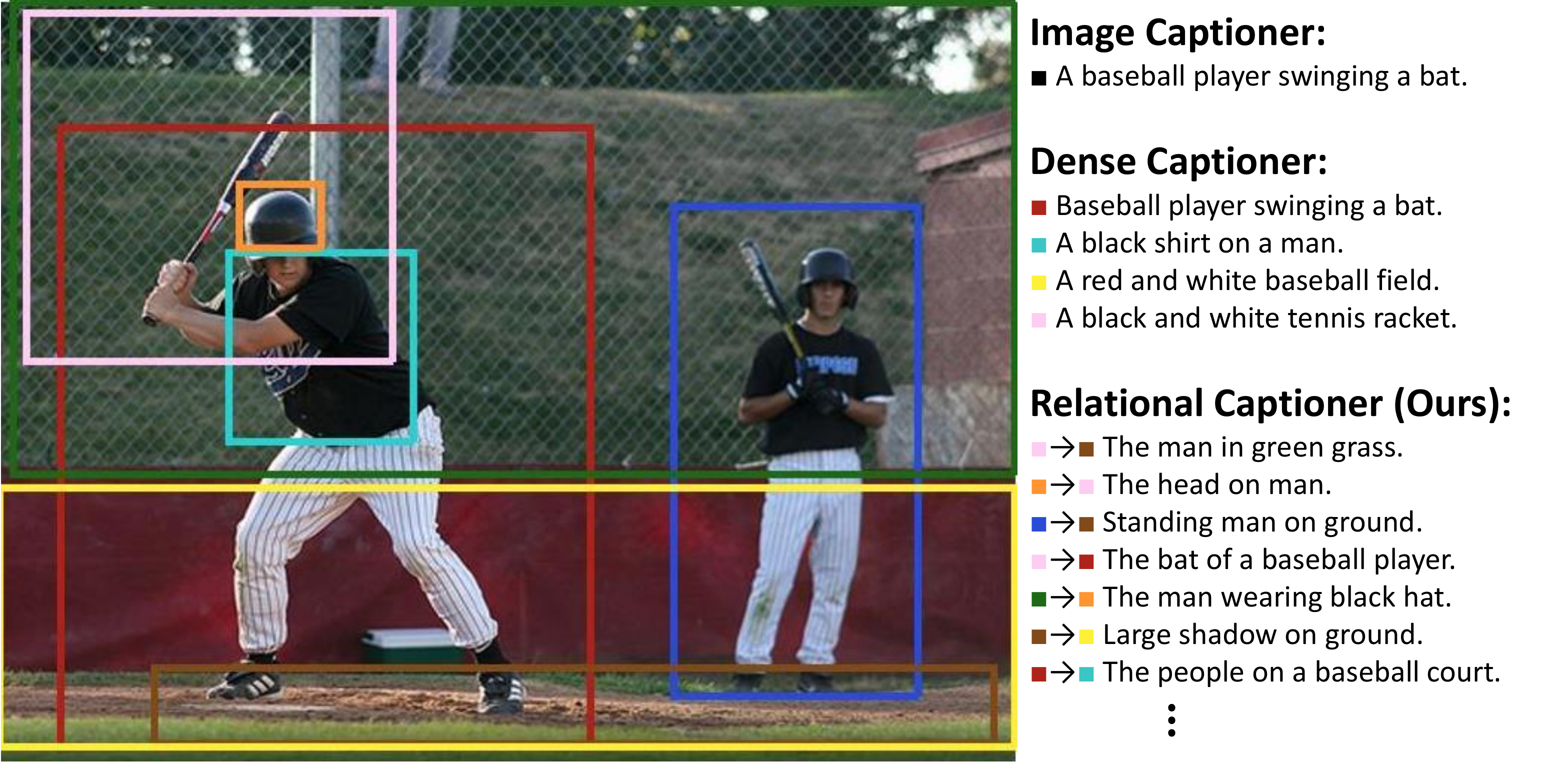}}\\
{\includegraphics[height=0.24\linewidth,keepaspectratio]{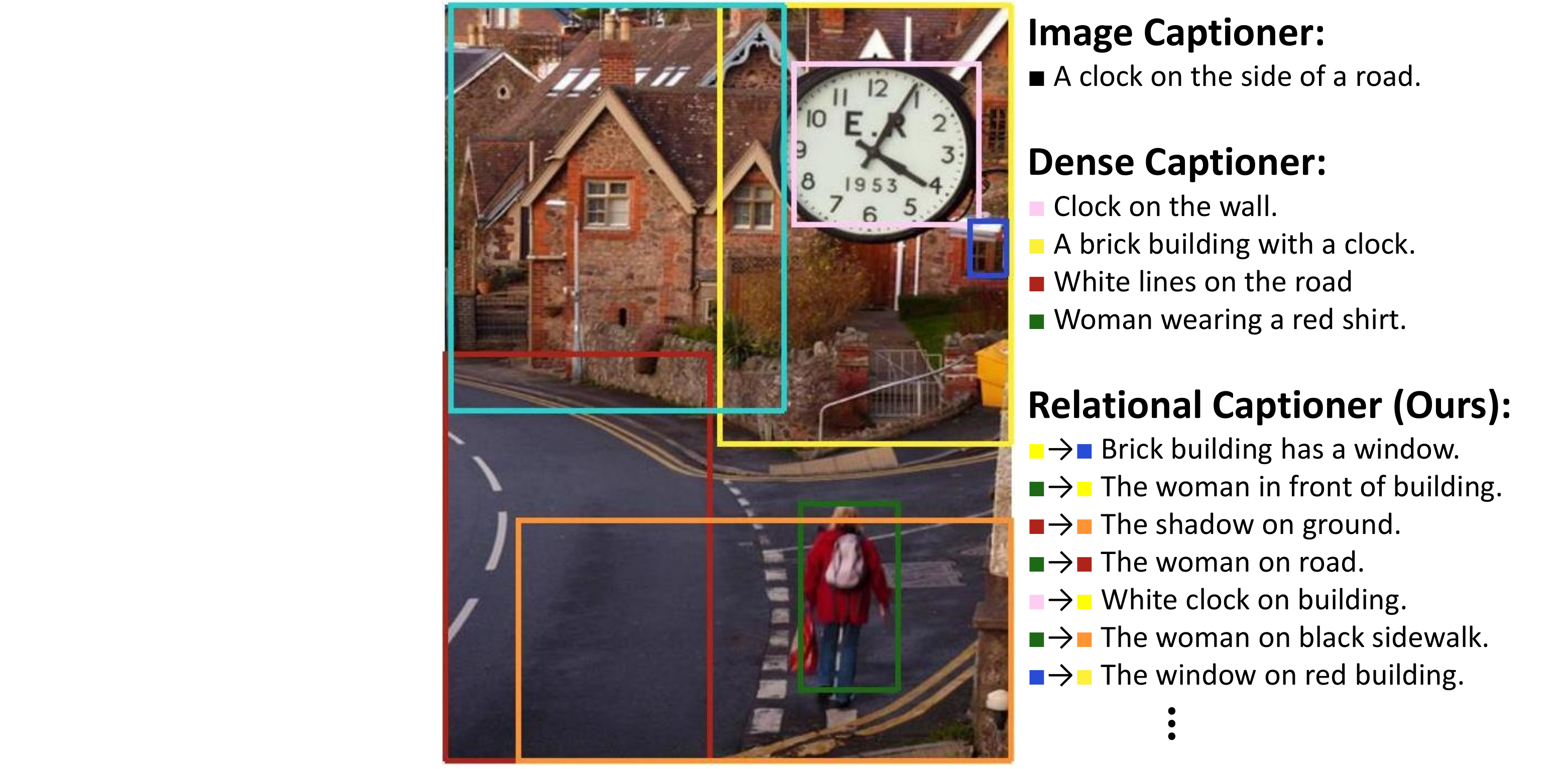}}
{\includegraphics[height=0.24\linewidth,keepaspectratio]{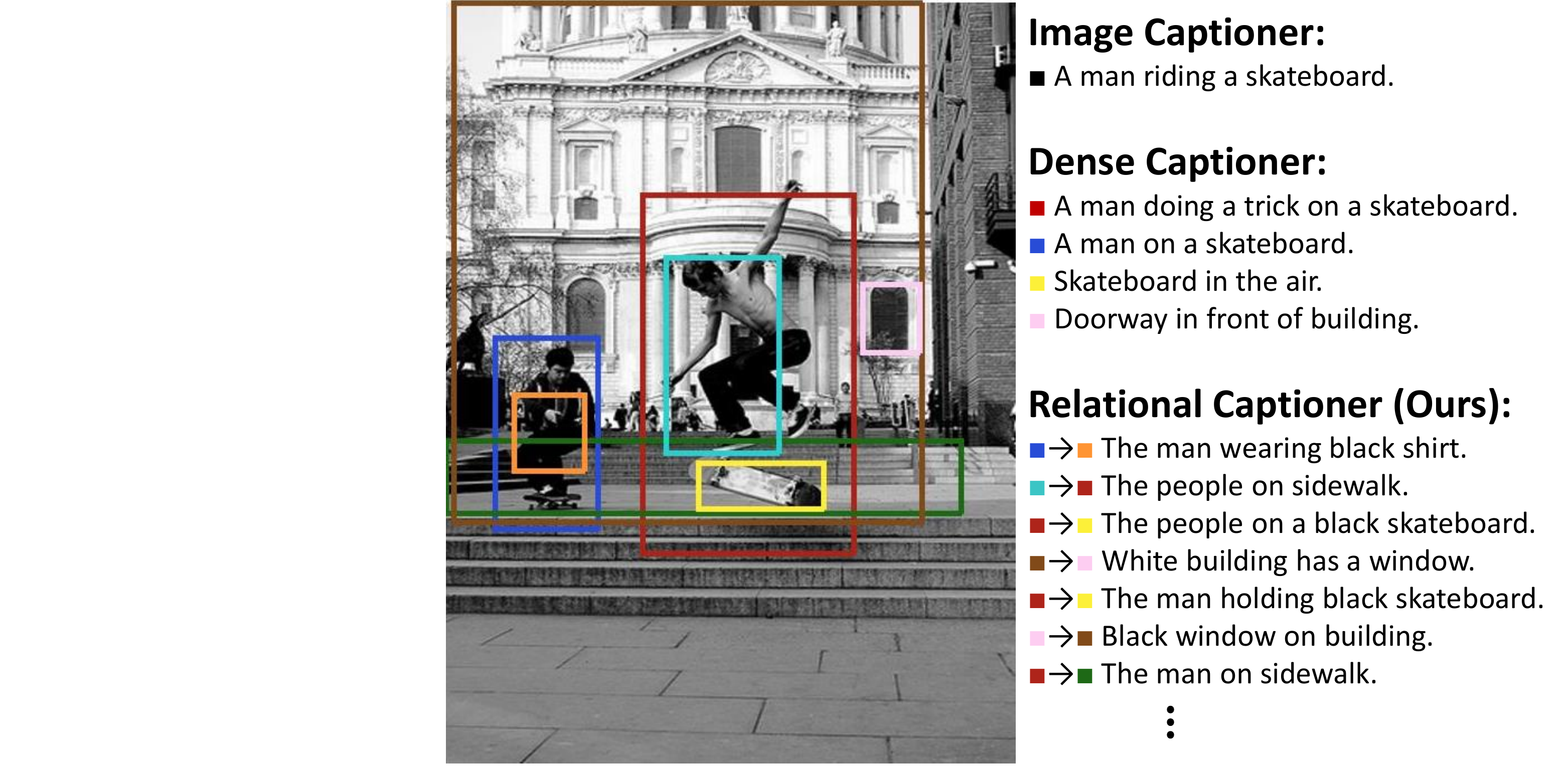}}\\
\vspace{-2mm}
\caption{Example captions and region generated by the proposed model on Visual Genome test images. 
The region detection and caption results are obtained by the proposed model from Visual Genome test images. 
   We compare our result with the image captioner~\cite{vinyals2015show} and the dense captioner~\cite{johnson2016densecap} in order to contrast the amount of information and diversity.
   \vspace{-2mm}}
\label{fig:qualitative}
\end{figure*}

\noindent\textbf{Evaluation metrics.}
Motivated by the evaluation metric suggested for the dense captioning task by Johnson~\etal\cite{johnson2016densecap},
we suggest a modified evaluation metric for the relational dense captioning.
{
Firstly, to assess the caption quality, we measure the average METEOR score~\cite{denkowski2014meteor} for predicted captions (noted as METEOR). 
Also, we use a mean Average Precision (mAP) similar to  Johnson~\etal which measures both localization and language accuracy.
For language accuracy,
we measure METEOR score with thresholds $\{ 0, 0.05, 0.1 0.15, 0.2, 0.25\}$, 
and we use IOU thresholds $\{0.2, 0.3, 0.4, 0.5, 0.6\}$ for localization accuracy.%, and use the same thresholds with METEOR for language.
}
The AP values, obtained by all the pairwise combinations of language and localization thresholds, are averaged to get the final mAP score.
The major difference of our metric from that of Johnson~\etal is that, for the localization AP, we measure for both the subject and object bounding boxes with respective ground truths. 
In particular, we only consider the samples with IOUs of both the subject and object bounding boxes greater than the localization threshold, which yields a \emph{more challenging metric}.
%Specifically, if the IOU of both subject and object bounding boxes are greater than a given localization threshold, this subject-object box pairs are regarded as ``positive,'' and captions that score greater than the language threshold as ``positive.'' 
%\ch{Pairs must satisfy both the localization and language thresholds to be regarded as ``positive'' samples which is accounted for measuring mAP.}
%\oh{Couldn't understand $\Rightarrow$ Then, the mAP score is measured by computing the ratio of true positives and the number of positive pairs. What is the mother set???}
%\newcommand{\djkim}[1]{\textcolor{magenta}{#1}}
For all cases, we use percentage as the unit of metric.

In addition, we suggest another metric, called ``image-level (Img-Lv.) recall.'' 
This measures the caption quality at the holistic image level by considering the bag of all captions generated from an image as a single prediction.
This metric evaluates the diversity of the produced representations by the model for a given image.
Specifically, with the aforementioned language thresholds of METEOR, we measure the recall of the predicted captions over about 20 ground truth captions.

\begin{figure*}[t]
\vspace{-2mm}
\centering
\resizebox{1\linewidth}{!}{%
% \footnotesize
    \begin{tabular}[c]{c}
    \subfigure[]{\includegraphics[width=0.35\linewidth,keepaspectratio]{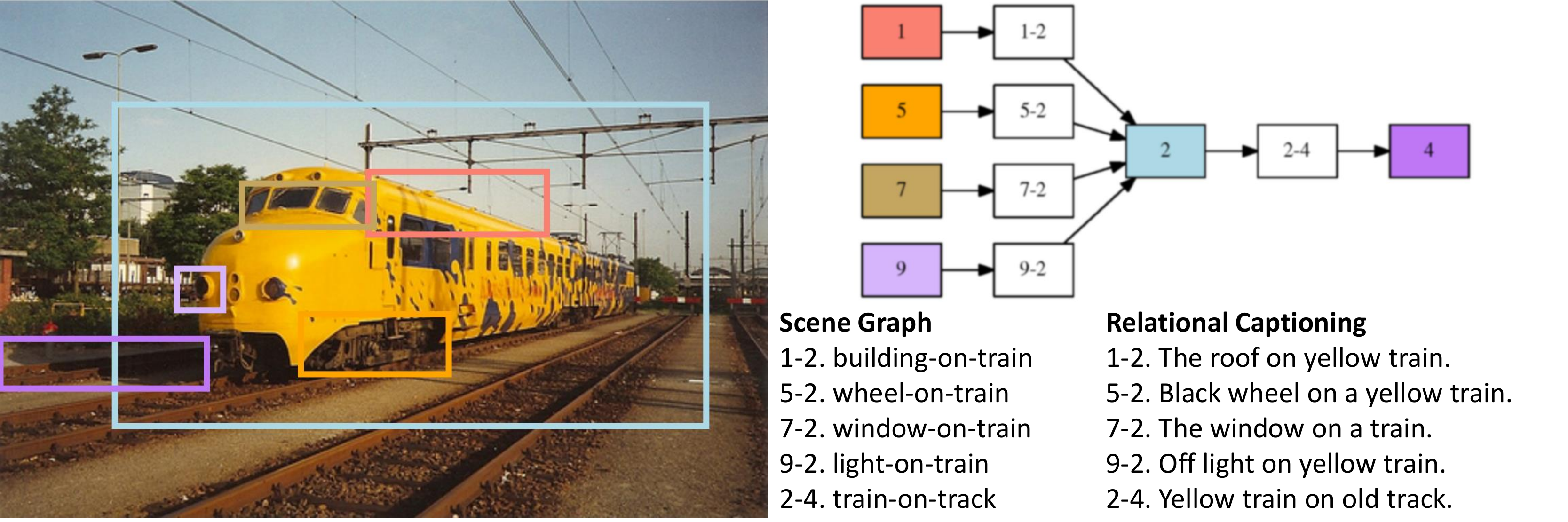}}				
    \subfigure[]{\includegraphics[width=0.35\linewidth,keepaspectratio]{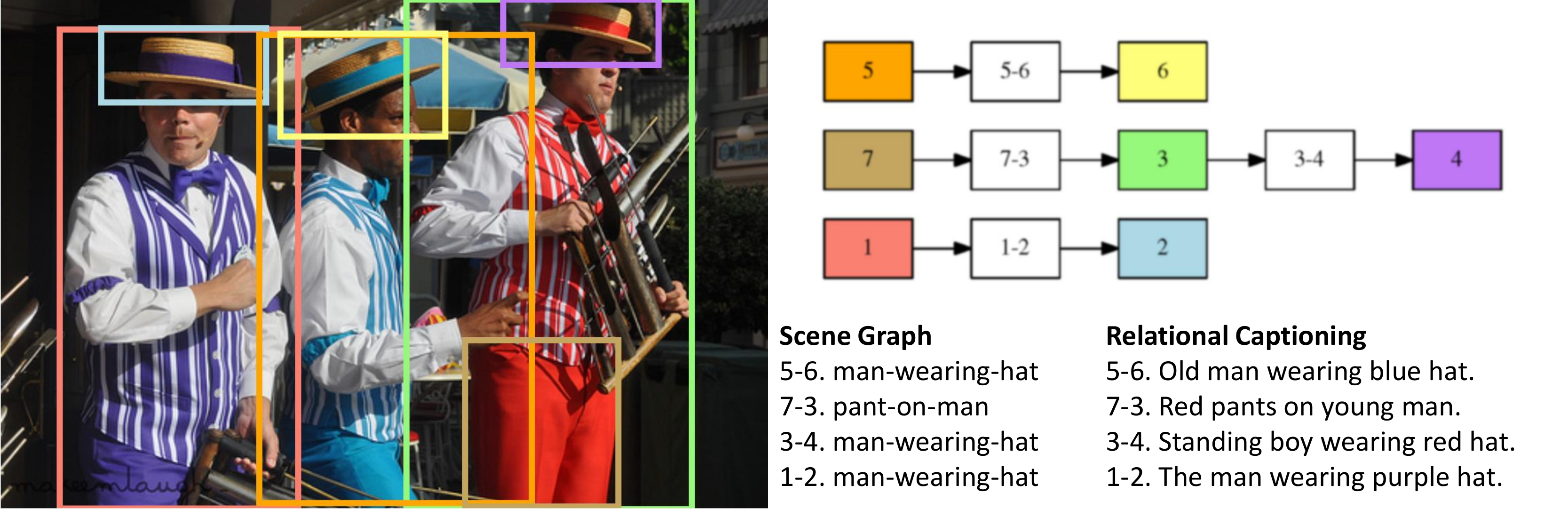}}\\
    \subfigure[]{\includegraphics[width=0.35\linewidth,keepaspectratio]{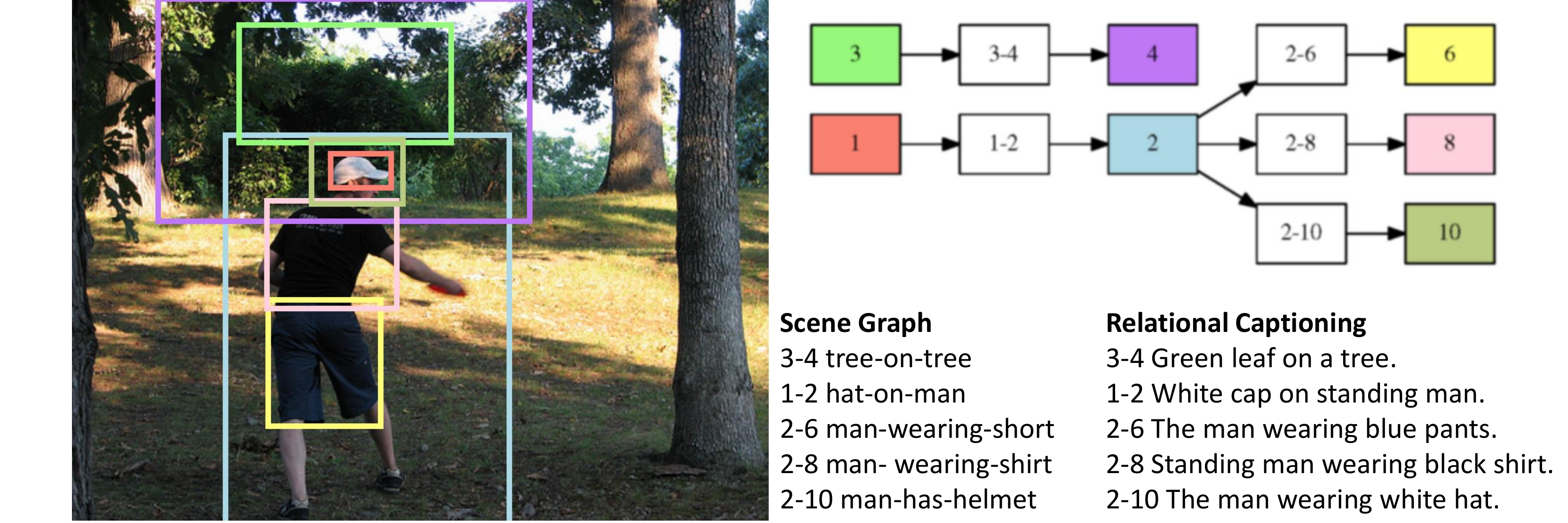}}
    \subfigure[]{\includegraphics[width=0.35\linewidth,keepaspectratio]{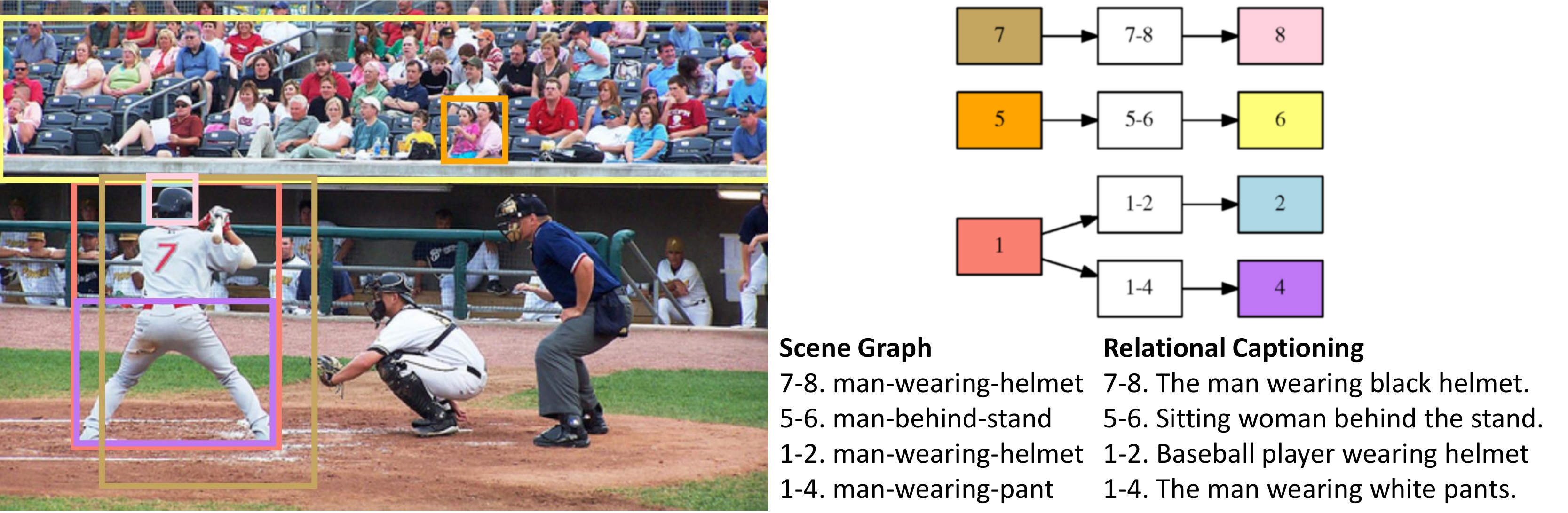}}\\
    \end{tabular}
}
	\vspace{-4mm}
	\caption{ Results of generating ``caption graph'' from our relational captioniner. In order to compare the diversity of the outputs, we also show the result of the scene graph generator, \texttt{Neural Motifs}~\cite{zellers2018neural}.\vspace{-2mm}} 
	\label{fig:scene-graph}
\end{figure*}

\begin{figure}[b]
\vspace{-4mm}
\centering
 \resizebox{1.0\linewidth}{!}{%
  {
  \subfigure[]{\includegraphics[width=0.3\linewidth]{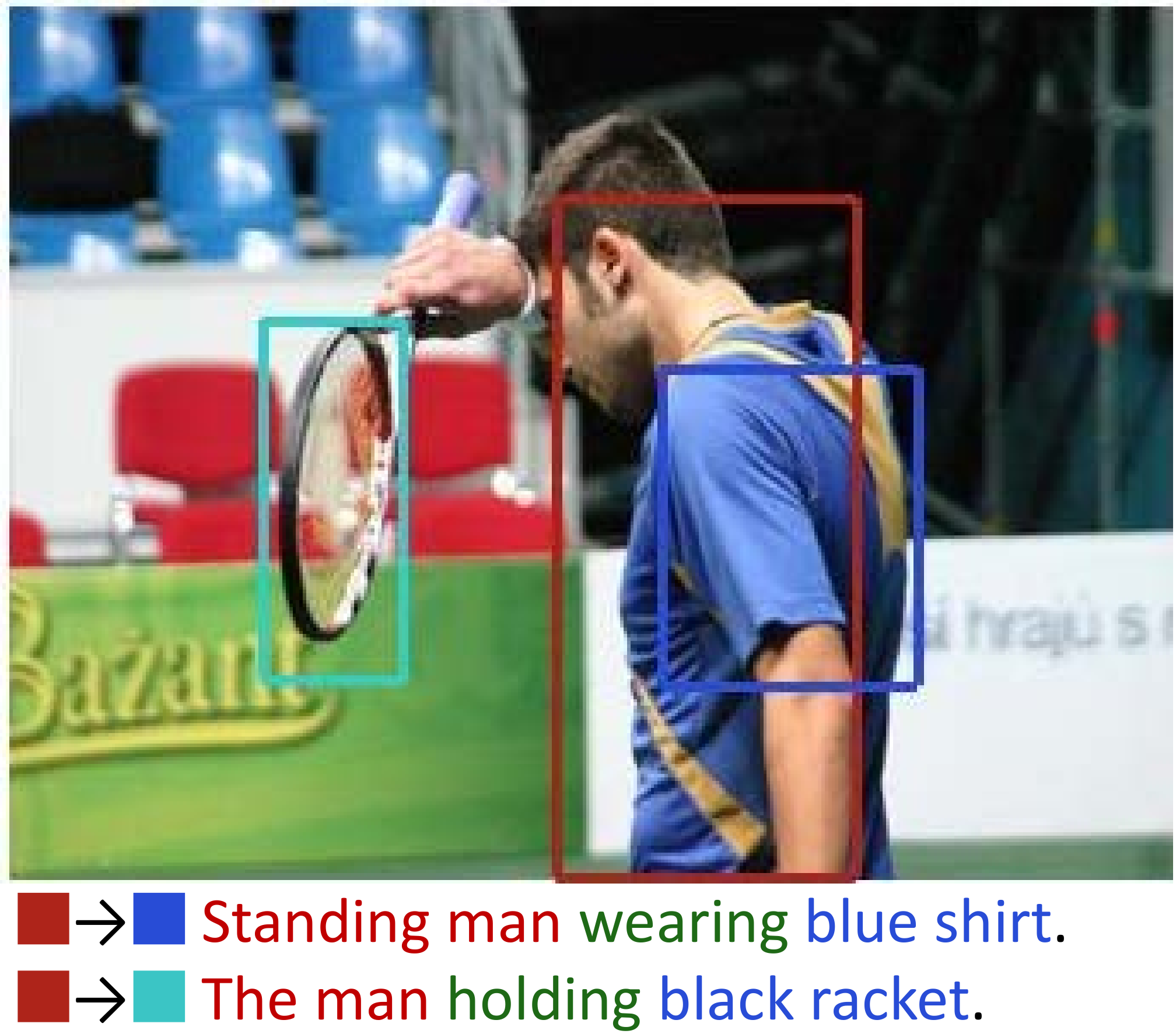}}
  \subfigure[]{\includegraphics[width=0.3\linewidth]{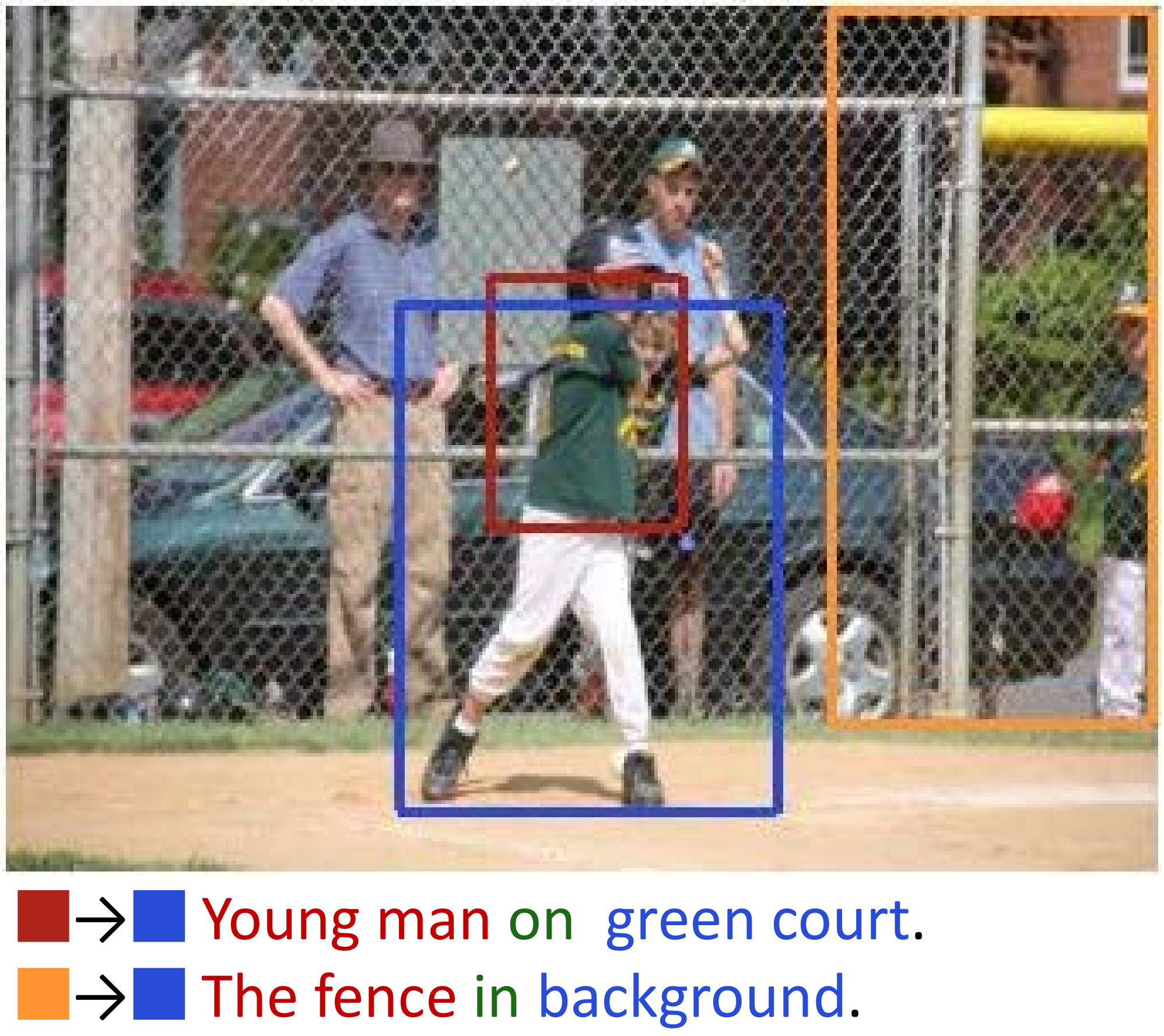}}
   \subfigure[]{\includegraphics[width=0.3\linewidth]{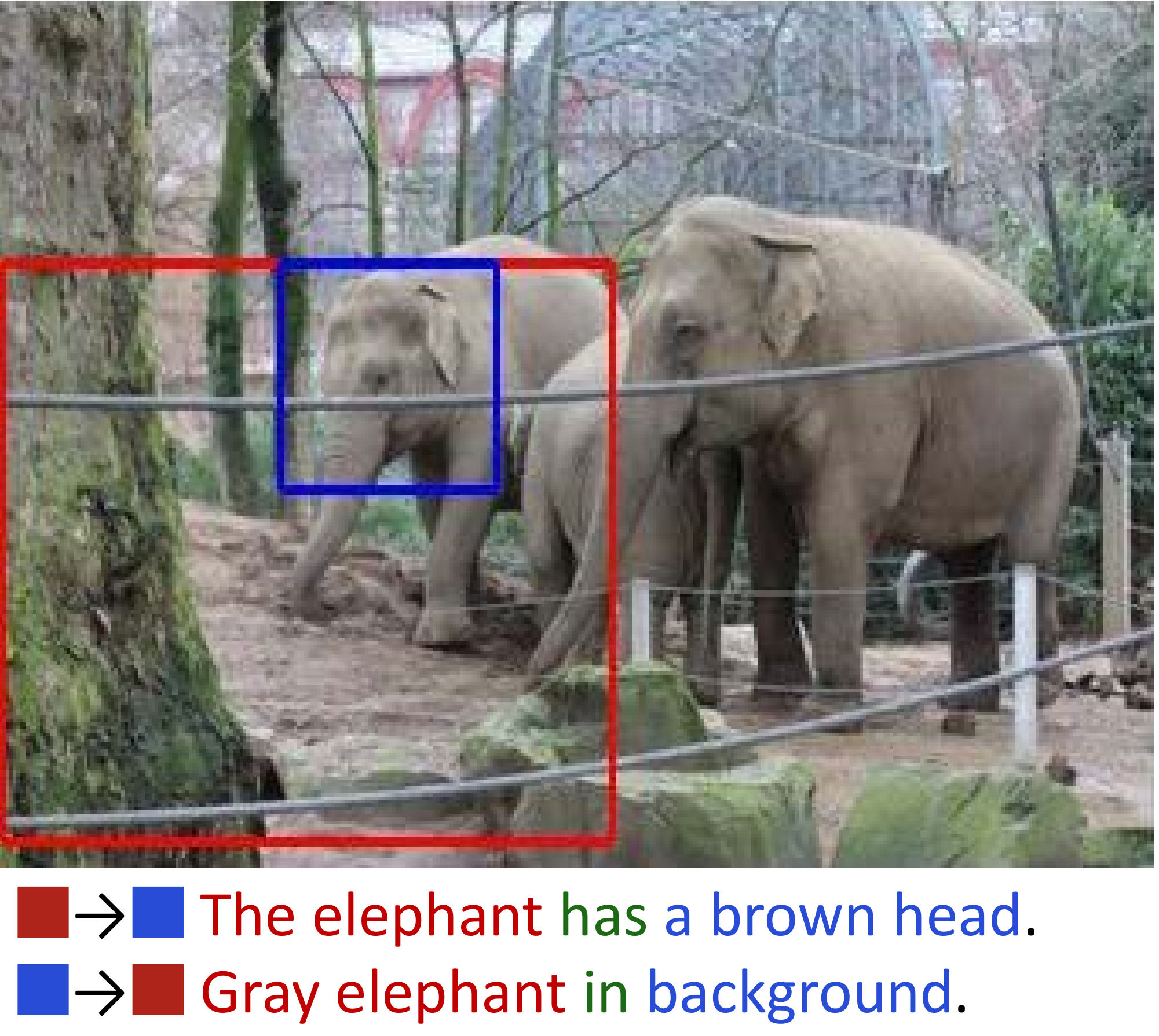}}
   }
  }
  \vspace{-4mm}
   \caption{Examples of different captions predicted from relational captioning by (a) changing objects, (b) changing subjects, and (c) switching the subject and object. Our model shows different predictions from different subject and object pairs.\vspace{-0mm}}
\label{fig:multibox}
\end{figure}

\noindent\textbf{Results.}
\Tref{table:captioning} compares the performance of {various methods for} the relational dense captioning task on {the} relational captioning dataset. 
{To compare with a different representation of relationship,} we additionally compare with the state-of-the-art scene graph generator, \texttt{Neural Motifs}~\cite{zellers2018neural}.
Due to the different output structure, we compare with \texttt{Neural Motifs} trained with the supervision for relationship detection.
Similar to the setup in \cite{johnson2016densecap}, we fix the number of region proposals after NMS to 50 for all methods for a fair comparison.

Within the second row section (2-7th rows) of \Tref{table:captioning}, our \texttt{TSNet} shows the best result suggesting that the triple-stream component alone is a sufficiently strong baseline over the others.
On top of \texttt{TSNet}, applying the MTL loss (\ie, \texttt{MTTSNet}) improves overall performance, and especially improves mAP, where the detection accuracy is
% seems to be 
dominantly improved compared to 
% the improvement of 
the other metrics.
This shows that \emph{triple-stream LSTM} is the key module that most leverages the MTL loss across other early fusion approaches (see the third {row section} of the table).
{Also, compared to \texttt{Union+Union+Union~(w/MTL)}, our \texttt{MTTSNet} shows much higher performance, which validates that the performance improvement by our method is not simply due to the increased number of the model parameters.}
Moreover, by adding REM to our late fusion method, \texttt{MTTSNet}, we have achieved further improvements in both mAP and Img-Lv. 
Recall scores (more strongly on Img-Lv. Recall).
As another factor, {we can see from \Tref{table:captioning} that the relative spatial information (\texttt{Coord.}) and union feature information (\texttt{Union}) improves the results.} 
%relationship learning in many cases, so does the union feature information (\texttt{Subj+Obj+Union}).
% Moreover, it seems that the union feature information acts similar to the spatial information in \texttt{Subj+Obj+Union}.
This is because 
the union feature itself preserves the spatial information to some extent from the $7\times 7$ grid form of its activation. 
% For \texttt{Neural Motifs}, other
Also, the relational captioner baselines including our \texttt{TSNet} and \texttt{MTTSNet} perform favorably against \texttt{Neural Motifs} in all metrics.
% \djkim{Several baselines including our models (\texttt{TSNet} and \texttt{MTTSNet}) even show better performance in all metrics.}
%This is worth noting because 
Note that handling free-form language generation which we aim to achieve is more challenging than the simple triplet prediction of scene graph generation.
% However, triple-stream networks show better performance than \texttt{Neural Motifs} in all metrics with a large margin.

% This is notable because 
%For reference, we also report a score for {Neural Motifs tested on VG relationship dataset (with $\dagger$).

\begin{table}[b]
	\centering 
    \resizebox{0.9\linewidth}{!}{%
    \begin{tabular}{l cccc}\toprule
    &		R@1		&	R@5	&	R@10		& 	Med	\\\midrule
    Image Cap. (\texttt{Full Image RNN})~\cite{karpathy2015deep}	    &		0.09	&	0.27	&	0.36	&	14	\\
    Dense Cap. (\texttt{Region RNN})~\cite{girshick2015fast}          &		0.19	&	0.47	&	0.64	&	6	\\
    Dense Cap. (\texttt{DenseCap})~\cite{johnson2016densecap}		    &		0.25	&	0.48	&	0.61	&	6	\\
    Dense Cap. (\texttt{TLSTM})~\cite{Yang_2017_CVPR}		&		0.27	&	0.52	&	0.67    &	5	\\
    \midrule
    Relational Cap. (\textbf{\texttt{MTTSNet}})		&{0.29} &{0.60}	&{0.73}&{4}\\
    Relational Cap. (\textbf{\texttt{MTTSNet+REM}})		&\textbf{0.32} &\textbf{0.64}	&\textbf{0.79}&\textbf{3}\\\midrule
    Random chance & 0.001 & 0.005	& 0.01 & - \\\bottomrule
    \end{tabular}    }
    \vspace{1mm}
	\caption{Sentence based image retrieval performance {comparison across different representations.}
% 	compared to previous frameworks. 
	We evaluate ranking using recall at $k$ ($R@K$, higher is better) and the median rank of the target image (Med, lower is better). 
	{The random chance performance 
	%of $R@K$ 
	is  provided for reference.
	{We compare with \texttt{TLSTM} in addition to the baselines (\texttt{Full Image RNN}, \texttt{Region RNN}, \texttt{DenseCap}) suggested in Johnson~\etal\cite{johnson2016densecap}.}
	}
	\vspace{-0mm}}
    \label{table:retrieval}	
\end{table}

\begin{figure*}[t]
	\vspace{-2mm}
	\centering 
    %\raisebox{\dimexpr\ht\imagebox-\height}
    \includegraphics[width=1\linewidth,keepaspectratio]{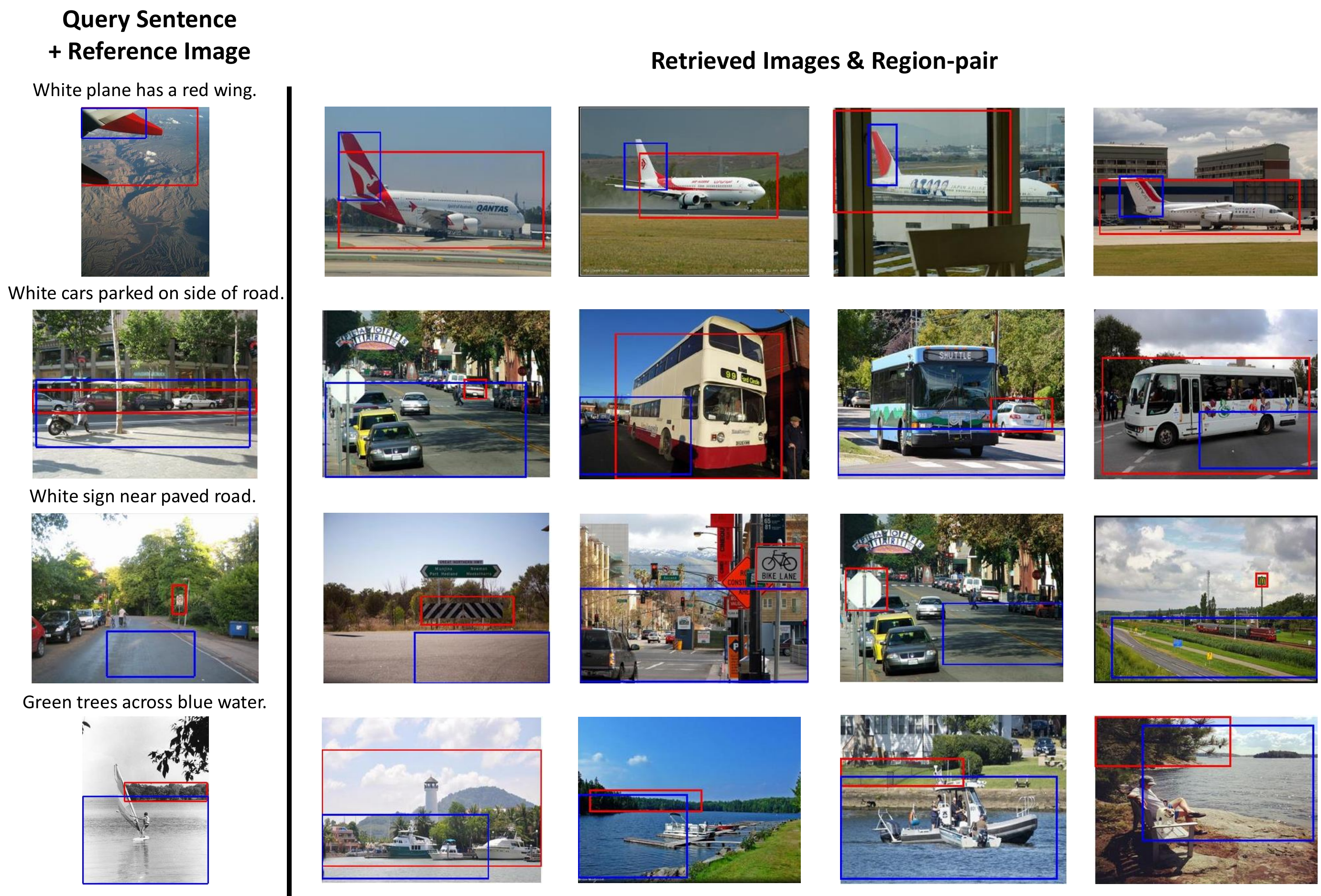}
    \vspace{-6mm}
	\caption{Sentence based image and region-pair retrieval results on Visual Genome test images. The retrieved results are shown in the {ranked} order.
    \vspace{-0mm}}
	\label{fig:retrieval}
\end{figure*}

%The baseline model is not able to distinguish the change of subject and object. Also, even if the union region is the same, it can different caption should be available based on the subject and object. \ch{(revision needed)}

%-----------------------4.3--------------------------------------------------
\subsection{{Comparison with} Holistic Image Captioning}
We also compare our approach with other image captioning frameworks, \emph{Image Captioner} {(\texttt{Show\&Tell}~\cite{vinyals2015show}, \texttt{SCST}~\cite{rennie2017self}, and \texttt{RFNet}~\cite{jiang2018recurrent}}), and \emph{Dense Captioner}~(\texttt{DenseCap}~\cite{johnson2016densecap} {and \texttt{TLSTM}~\cite{Yang_2017_CVPR}}) in a holistic image description perspective.
To measure the performance of \emph{holistic image-level} captioning for dense captioning methods, we use Img-Lv. Recall metric {defined in the previous section}.
We compare them with two relational dense captioning methods, \texttt{Union} and \texttt{MTTSNet} ({as well as \texttt{+REM}}), denoted as \emph{Relational Captioner}.
For a fair comparison, for \emph{Dense} and \emph{Relational Captioner}, we adjust the number of region proposals after NMS to be similar, which is different from the setting in the previous section which fixes the number of proposals before NMS.
{For fair comparison with the \emph{Image Captioner}, in addition to  {the typical selection of words according to maximum probabilities in} 
% max probability sampling for 
caption generation,
% (selecting a word that has the highest probability from a model), 
we introduce another baselines using} a stochastic sampling (probabilistically selecting a word 
{proportional to}
% based on
{the probabilities of words} from a model) to 
{allow diverse caption generation from the LSTM.}
% generate multiple captions from a LSTM. 
We generate 10 captions from {the stochastic variant image captioners} {in order to match the number of captions between \emph{Image Captioner} and \emph{Dense Captioner}}.
% an image captioner. 
% The image captioners with this option are noted with $\dagger$.
% {We annotate such variant of image captioners with $\dagger$ symbol.}
%Specifically, we set the maximum number of boxes for NMS to be 85 for dense captioner and 75 for relational captioner.
%{We also show a result of the proposed network trained and evaluated on the dataset with the proposed augmentation method, denoted as \texttt{MTTSNet} w/ augmentation.}
{Finally, in order to isolate the performance of the caption generation and the box localization modules, we measure the captioning performance by setting the bounding boxes as the ground truth boxes. 
We annotate such variant of relational captioners with $(GT)$.}

\Tref{table:recall} {compares} the image-level recall, METEOR, and additional quantities. 
\emph{\#Caption} denotes the average number of captions generated from an input image and \emph{Caption/Box} denotes the average ratio of the number of captions generated and the number of boxes remaining after NMS.
Therefore, \emph{Caption/Box} demonstrates how many captions can be generated given the same number of boxes generated after NMS.
By virtue of multiple captions per image from multiple boxes, the \emph{Dense Captioner} is able to achieve higher performance than {all  the} \emph{Image Captioner}s. 
{While {the} stochastic sampling {methods slightly improve} 
% slightly improved 
image captioning performance in terms of recall, 
% as the diversity that a single image captioner can have is very limited, 
the performance is still far lower than \emph{Dense Captioner}s or \emph{Relational Captioner}s by a large margin,
{as the diversity of an image captioner's output is still very limited by its inherent design.}
}
Compared with the \emph{Dense Captioner}s, \texttt{MTTSNet} as a \emph{Relational Captioner} can generate an even larger number of captions, given the same number of boxes.
Hence, as a result of learning to generate diverse captions, the \texttt{MTTSNet} achieves higher recall and METEOR.
{\texttt{TLSTM}~\cite{Yang_2017_CVPR} 
% can 
improves the performance 
% compared to 
of
\texttt{DenseCap}~\cite{johnson2016densecap} 
% by virtue of 
due to a better representational power, but the performance is still lower than that of \texttt{MTTSNet}.}
% From the performance of 
Comparing to
\texttt{Union}, we can see that it is difficult to obtain better captions than \emph{Dense Captioner} by only learning to use the union of subject and object boxes, despite having a larger number of captions.
{Adding \texttt{REM} {to our \texttt{MTTSNet},} further 
% improved 
improves
the performance in both the Recall and the METEOR score.}
{In addition, even when setting the bounding boxes as the ground truth bounding boxes, by virtue of the more powerful language module, \texttt{MTTSNet} (especially \texttt{MTTSNet+REM}) shows favorable performance compared to \texttt{Union}.}

We show prediction examples of our relational captioning model in \Fref{fig:qualitative} {along with the comparisons against the traditional frameworks, image captioner~\cite{vinyals2015show} and dense captioner~\cite{johnson2016densecap}.}
Our model is able to generate rich and diverse captions for an image, {compared to other paradigms.}
% We also show a comparison with .
While the dense captioner is able to generate diverse descriptions than an image captioner by virtue of {localized} regions, our model can generate an even more number of captions from the combination of the bounding boxes.
%Note that the image captioner~\cite{vinyals2015show} and dense captioner~\cite{johnson2016densecap} are trained on the ``captioning'' dataset, whereas our model is trained on the relationship dataset where we tokenized the labels \texttt{subj-pred-obj} into separate words.

%\subsection{Qualitative Results of Relational Captioning}
%\Fref{fig:qualitative} shows additional qualitative results on relational dense captioning task.
%Note that the image captioner~\cite{vinyals2015show} and dense captioner~\cite{johnson2016densecap} are trained on the ``captioning'' dataset, whereas our model is trained on the relationship dataset where we tokenized the labels \texttt{subj-pred-obj} into separate words.
\Fref{fig:multibox} shows caption prediction examples for multiple box pair combinations. 
{Based on the output of {the} POS predictor, we color the words of the caption as (red, green, blue) for (\texttt{subj}-\texttt{pred}-\texttt{obj}) respectively.}
We note that, while the traditional dense captioning simply takes a single region as input and predicts one dominant description, in 
% the proposed relational captioning 
our framework, different captions can be obtained from different subject and object pairs. 
In addition, one can see that the predicted POS is correctly aligned with the words in the generated captions. 
% Moreover, as 
Although the POS classification is not our target task, for completeness, we measure the accuracy of the \texttt{MTTSNet} POS estimation by comparing it with 
% the POS prediction from \texttt{MTTSNet} with
the ground truth POS, which 
% of the ground truth caption, 
% The accuracy 
is 89.7\%.
%This cannot be seen with VRD or scene graph models.
{The detailed accuracies for subject, predicate, and object are 91.6\%, 86.5\%, and 90.9\%, respectively.}

\begin{figure*}[t] 
	\vspace{-2mm}
	\centering
	\includegraphics[width=1.0\linewidth,keepaspectratio]{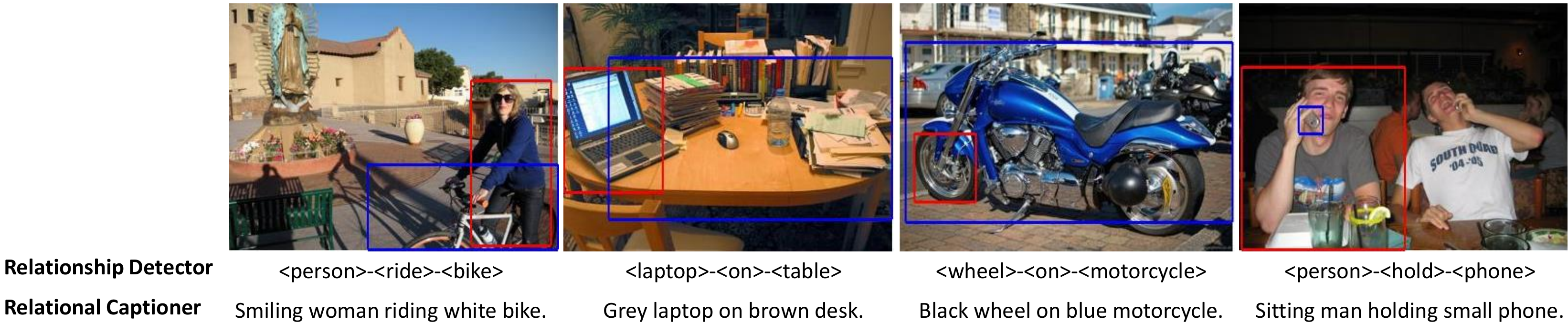}
	\caption{Qualitative comparison with visual relationship detection model~\cite{lu2016visual}. The proposed relational captioning model is able to provide more detailed information than the traditional relationship detection model.\vspace{-0mm}} 
	\label{fig:VRD_comparison}
\end{figure*}

\begin{table}[t]
\resizebox{1.0\linewidth}{!}{%
\begin{tabular}{l ccc}\toprule
&	mAP (\%)	&	Img-Lv. Recall 	& METEOR\\\midrule

\texttt{Direct Union}&	--	&	54.51	&25.53\\\midrule
\texttt{Union} &	1.66	& 	54.30	&24.82\\	
%union+coordinate				&\textbf{2.20(12.42)}	& 	3.12(16.01)	&62.84(31.62)\\			
\texttt{Union+Coord.}				&{1.90}	& 	64.11		&30.81\\	
\texttt{Subj+Obj}					&1.90		& 	55.06	& 25.09\\
\texttt{Subj+Obj+Coord.}				&1.68	& 	68.33	& 33.45\\
\texttt{Subj+Obj+Union}			&{1.94}	& 	{68.32}	& {33.77}\\
%\texttt{Subj+Obj+Union}	(new)				&{1.94}	& 	69.78	& 35.12\\
\texttt{\textbf{TSNet (Ours)} }		&\textbf{1.99}	&	\textbf{68.44} 		&\textbf{34.49}\\
\midrule
\texttt{Union (w/MTL)}					&1.70		& 	66.39	&31.62\\
\texttt{Subj+Obj+Coord (w/MTL)}		&	1.93 	&	68.80 	&33.49\\
\texttt{Subj+Obj+Union (w/MTL)}		& {2.17}	& 	65.04	& 32.25\\
\texttt{\textbf{MTTSNet (Ours)}}			&{2.18}& 	{71.44}	&{35.47}\\
\texttt{\textbf{MTTSNet (Ours)+REM}}			&\textbf{2.21}& 	\textbf{73.36}	&\textbf{35.65}\\
\midrule
\texttt{\textbf{MTTSNet (Ours)+REM (R)}}			&{2.33}& 	{77.44}	&{37.63}\\
\midrule
\texttt{Language Prior}~\cite{lu2016visual} &2.13 & 46.60&28.12 \\
\texttt{Shuffle-Then-Assemble}~\cite{yang2018shuffle}& 2.20 & 69.98&29.50 \\
\bottomrule
\end{tabular}
}
\vspace{0mm}
\caption{{Ablation study on the relational dense captioning task with the VRD dataset. {Our \texttt{TSNet} and \texttt{MTTSNet} ({both with and without \texttt{+REM}}) show top
{performance} among the relational captioning models.
% with and without POS prediction (w/MTL) respectively. 
{(R) indicates ResNet-50~\cite{he2016deep} as a backbone network instead of VGG-16.}
% \djkim{(R) indicates the results when using ResNet-50~\cite{he2016deep} instead of VGG-16.}
In addition, \texttt{MTTSNet} ({both with and without \texttt{+REM}}) shows favorable performance against {the} VRD models~\cite{lu2016visual,yang2018shuffle} with a noticeable
% large
margin.}} \vspace{-6mm} }
\label{table:ablation_VRD}
\end{table}

%-----------------------4.4--------------------------------------------------
\subsection{Comparison with Scene Graph}
{
Motivated by scene graph, which is derived from the VRD task,
we extend to a new type of a scene graph, which we call ``caption graph.'' 
\Fref{fig:scene-graph} shows the caption graphs generated from our \texttt{MTTSNet} as well as the scene graphs from \texttt{Neural Motifs}~\cite{zellers2018neural}.
{For caption graph, we follow the same procedure with \texttt{Neural Motifs}, but replace the relationship detection network {with} our \texttt{MTTSNet}.}
In both methods, we use ground truth bounding boxes 
%{to generate scene (and caption) graphs} 
for fair comparison.

By virtue of being free form, our caption graph can have richer expression and information including attributes, whereas the traditional scene graph is limited to a closed set of the \texttt{subj-pred-obj} triplet.
For example, in \Fref{fig:scene-graph}-(b,d), given the same object ``person,'' our model is able to distinguish the fine-grained category (\ie, man vs boy and man vs woman).
In addition, our model can provide more status information about the object (\eg, standing, black), by virtue of the attribute contained in our relational captioning data.
Most importantly, the scene graph can contain unnatural relationships (\eg, tree-on-tree in \Fref{fig:scene-graph}-(c)), because the back-end relationship detection methods, \eg, \cite{zellers2018neural}, predict object classes independently.
In contrast, by predicting the full sentence for every object pair, {the} relational captioner can assign a more appropriate word with attributes for an object by considering the relations, \eg, ``Green leaf on a tree.''

Lastly, our model is able to assign different words for the same object by considering the context (the man vs baseball player in \Fref{fig:scene-graph}-(d)), whereas the scene graph generator can only assign one most likely class (man).
Thus, our relational captioning framework enables more diverse interpretation of the objects compared to the traditional scene graph generation models, which would be more favorable representation to scene context understanding.
%For comparison with VRD task models, we show both qualitative and quantitative comparisons in the supplementary material.
}

\begin{figure*}[t]
	\vspace{-2mm}
	\centering
	\includegraphics[width=.8\linewidth,keepaspectratio]{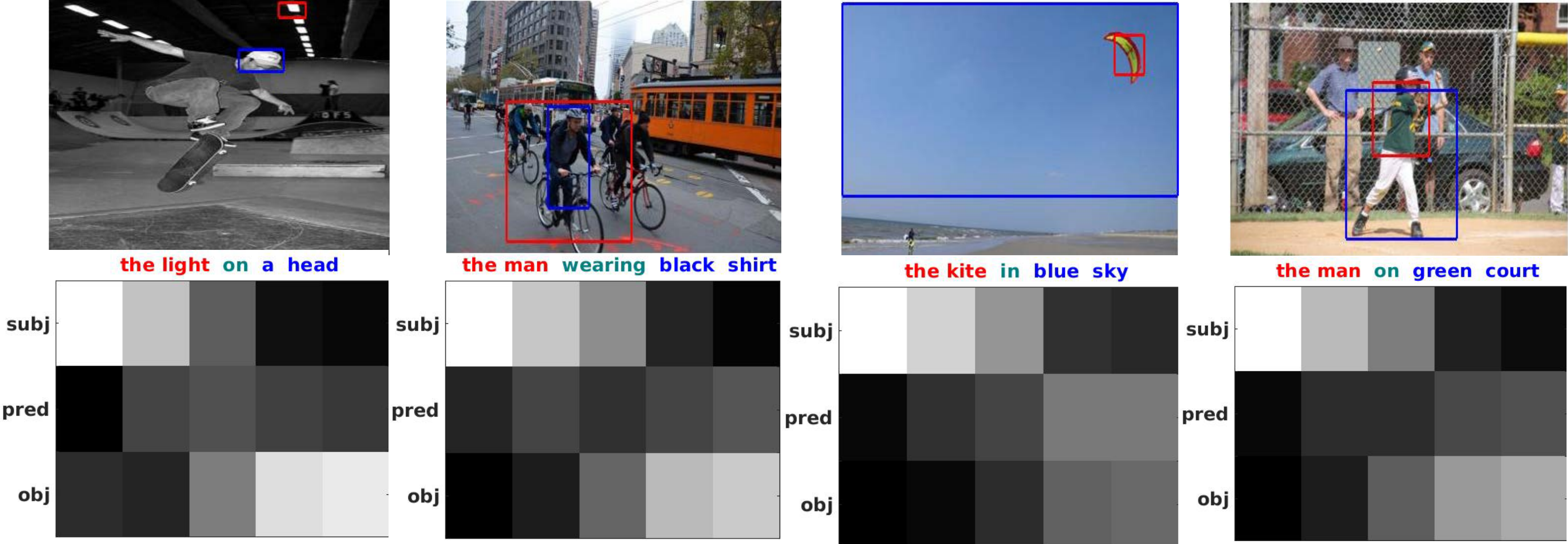}
    \vspace{-0mm}
	\caption{{Visualization of POS importance transition}. 
		Y-axis represents respective representative hidden values of \texttt{Subject}-\texttt{Predicate}-\texttt{Object} LSTMs, and X-axis represents {words of each caption in order.}
		{\texttt{subj}-\texttt{pred}-\texttt{obj} are color-coded by red, green, and blue colors according to the output of the POS predictor, respectively. {Each word in the captions comes from the corresponding LSTM.}}
		\vspace{-2mm}
% 		time steps in the caption.
		}
	\label{fig:weighttransfer}
\end{figure*}

\begin{table}[t]{%
\centering
    \resizebox{.9\linewidth}{!}{%
    \setlength{\tabcolsep}{1mm}
		\begin{tabular}{l ccc ccc}\toprule
			&\multicolumn{3}{c }{Phrase detection}&\multicolumn{3}{c}{Relationship detection}\\%\cline{2-7}
            &mAP&		R@100	&	R@50		&mAP&		R@100	&	R@50 	\\\midrule
            % 			language prior~\cite{lu2016visual} 	
            % \cite{lu2016visual_rebut}
            % {[Lu et al.]}
            \texttt{Language Prior}~\cite{lu2016visual} 	&{2.07}&		17.03&16.17&{1.52}&14.70&13.86		\\
            %             ViP-CNN~\cite{li2017vip}	% \cite{li2017vip_rebut}	

             \texttt{VTransE}~\cite{zhang2017visual} &-&22.42&19.42&-&15.20&14.07		\\
             \texttt{VRL}~\cite{liang2017deep}       &-&22.60&21.37&-&20.79&18.19\\
             \texttt{ViP-CNN}~\cite{li2017vip}	    &-&27.91&22.78&-&20.01&17.32\\
             \texttt{DR-Net}~\cite{dai2017detecting} &-&23.45&19.93&-&20.88&17.73	\\
             
             \texttt{CAI}~\cite{zhuang2017towards}   &-&19.24&17.60&-&17.39&15.63\\
             \texttt{PPR-FCN}~\cite{zhang2017ppr}    &-&23.15&19.62&-&15.72&14.41	\\
             Yu~\etal~\cite{yu2017visual}   &-&24.03&23.14&-&21.34&19.17		\\
             \midrule
             %Zoom-Net~\cite{yin2018zoom}    &-&28.89&25.21&-&22.39&19.54\\
            \textbf{\texttt{MTTSNet (Ours)}}		&{2.88}  & 20.98& {20.64} & {1.59} & 20.05 & {17.49} \\
            \textbf{\texttt{MTTSNet (Ours)+REM}}		&{2.91}  & 21.54& 21.39 & 1.64 & 20.70 & 17.74 \\
            %\midrule
            \textbf{\texttt{MTTSNet (Ours)+REM (R)}}		&3.09  & 28.40& 24.18 & 1.73 & 21.87 & 19.36 \\
        \bottomrule
		\end{tabular}
    }
    \vspace{1mm}
 	\caption{Comparison of our \texttt{MTTSNet} with VRD models {on the} VRD metrics on the VRD dataset. 
 	{(R) indicates ResNet-50~\cite{he2016deep} as a backbone network instead of VGG-16.}
 	{Despite the disadvantages for predicting complex captions compared to simple triplets, our \texttt{MTTSNet} and \texttt{MTTSNet+REM} show favorable {or comparable} performance against {the} VRD models.
 	{Also, note that most of the VRD models have the benefit of strong backbones such as ResNet, but our \texttt{MTTSNet+REM (R)} with ResNet-50 surpasses all the other competing methods even with the VRD metrics unfavorable to ours.}
%  	\djkim{resnet version added. surpassing various VRD methods}
 	}\vspace{-6mm}
 	}
 	\label{table:VRDmetric}	
}
\end{table}

%-----------------------4.5--------------------------------------------------
\subsection{Sentence-based Image and Region-pair Retrieval}
Since our relational captioning framework produces richer image representations than other frameworks, it may have benefits on image {and}
% or
region-pair retrieval by sentence.
{Our method can directly deal with free-form natural language queries, whereas scene graph or VRD models require additional processing to handle the free-form queries.} %\emph{which {cannot} be performed by scene graph or VRD models \djkim{because these frameworks cannot handle free-form natural language queries}}. 
{In this section,} we evaluate our method on the retrieval task.
Following the same procedure 
%suggested by Johnson~\etal
from \cite{johnson2016densecap} but with our relational captioning dataset,
we randomly choose 1,000 images from the test set, and from these chosen images, we collect 100 query sentences by sampling four random captions from 25 randomly chosen images.
The task is to retrieve the correct image for each query by matching it with the generated captions.

{Our relational captioning based retrieval is done as follows.}
% Our matching score is computed as follows.
% In order to compute the matching score between the sentence and the image for the proposed model,
% we follow the same process suggested by John\cite{johnson2016densecap}. 
For every test image, we generate 100 region proposals from the RPN followed by NMS.
{To measure the degree of association, \ie, matching score,}
% In order to produce a matching score 
between a query and a region pair in the image, we compute the probability that the query text may occur from the region pair by multiplying the probability of words over recursive steps.
% would like to deduce the probability that the query caption is generated from an image.
Among all the scores {of} the region pairs from the image, we take the maximum matching score value as a representative score of 
{matching between the query text and} the image.
% This score is used as the matching score between the query text and the image; 
The retrieved images are sorted {according to} these computed matching scores.

{
We compare the retrieval performance with several baselines in \Tref{table:retrieval}.
We measure \emph{recall at top K}, \emph{R@K}, which is the success ratio across all the queries that, by each given query,
its ground-truth image is retrieved within top $K$ ranks.
We report $K \in \{1,5,10\}$ cases.
}
% We compute the ratio of the number of queries, of which the retrieved image ranked within top $k \in \{1,5,10\}$, and the total number of queries (denoted as ).
We also report the median rank of the correctly retrieved images across all 1000 test images.
{We follow the same procedure by Johnson~\etal of running through random test sets 3 times to report the average results.
% \djkim{In addition, we add the retrieval result with \oh{a} more powerful dense captioning model, \texttt{TLSTM}~\cite{Yang_2017_CVPR}.}
{We add an additional retrieval result with a more competitive dense captioning model, \texttt{TLSTM}~\cite{Yang_2017_CVPR}.}
From the result, our proposed relational captioners show favorable performance against the baselines.
% {This is meaningful because region pair based method is a more challenging setup than a single region based baseline approaches.}
{This is meaningful because a region pair based method deals with a more difficult input form than that of the single region based approaches.}
}
{Moreover, \texttt{MTTSNet+REM} {consistently} shows better retrieval performance compared to \texttt{MTTSNet}.}
% For baseline models \emph{Full Image RNN}, \emph{Region RNN}, and \emph{DenseCap}, we display the performance measured from Johnson~\etal\cite{johnson2016densecap}.
% To be compatible, 

% As shown in \Tref{table:retrieval}, the proposed relational captioner outperforms all baseline frameworks. 

\begin{table}[b]
\vspace{-2mm}
\centering
    	\resizebox{.8\linewidth}{!}{%
% 		\begin{tabular}{l|| c|c|c}\hline
% 													&		total words		&	words/img		&	words/box	\\\hline\hline
% 			Image Cap.~\cite{vinyals2015show}	&		96		&		3.05		&	-	\\
%             Dense Cap.~\cite{johnson2016densecap}		&		209		&		9.51		&	3.02	\\
% 			%RelCap(union-only)						&		?		&		?		&	?	\\
% 			Relational Cap. (\textbf{\texttt{MTTSNet}})				&{170} &\textbf{11.20}	&\textbf{9.57}\\\hline
% 		\end{tabular}
        \begin{tabular}{l cc}\toprule
													&			words/img		&	words/box	\\\midrule
			Image Cap.~\cite{vinyals2015show}	&				4.16		&	-	\\
            Scene Graph~\cite{zellers2018neural}				&		7.66		&	3.29	\\
            Dense Cap.~\cite{johnson2016densecap}		&				18.41		&	4.59	\\
            Relational Cap. (\textbf{\texttt{MTTSNet}})	 			&{20.45}&{15.31}\\
            Relational Cap. (\textbf{\texttt{MTTSNet+REM}})	 			&\textbf{25.57}&\textbf{18.02}\\
            \bottomrule
		\end{tabular}
        }\vspace{1mm}
	\caption{Diversity comparison between image captioning, scene graph generation, dense captioning, and relational captioning {frameworks}.
    We measure the number of different words per image (words/img) and the number of words per bounding box (words/box).\vspace{-0mm}}
	\label{table:diversity}	
\end{table}

\Fref{fig:retrieval} shows the qualitative results on the sentence based {image and region-pair} retrieval.
Given a sentence query, we show the retrieved images and their region pairs
with
the maximum matching score.
Image retrieval based on our approach has a distinct advantage in that it retrieves images containing similar contextual relationships despite significant visual differences.
More specifically, in the 3rd row of \Fref{fig:retrieval}, our method can retrieve images with an abstract contextual relationship of ``White sign near paved road.'' 
{The retrieved images} are visually diverse but share the same contextual information.
{Also, the natural language based retrieval from our framework is distinctive compared to traditional relationship detection methods (classification) which cannot handle natural language queries with variable length due to {their}
fixed form {input} (\ie\texttt{subj-pred-obj}).}
For example, in the 1st row, given a query that specifies the color ``red,'' our model is able to retrieve images of a plane with red wings \emph{which VRD models are not capable of}.

\begin{figure}[b]
\vspace{-2mm}
\centering
	\includegraphics[width=1.0\linewidth,keepaspectratio]{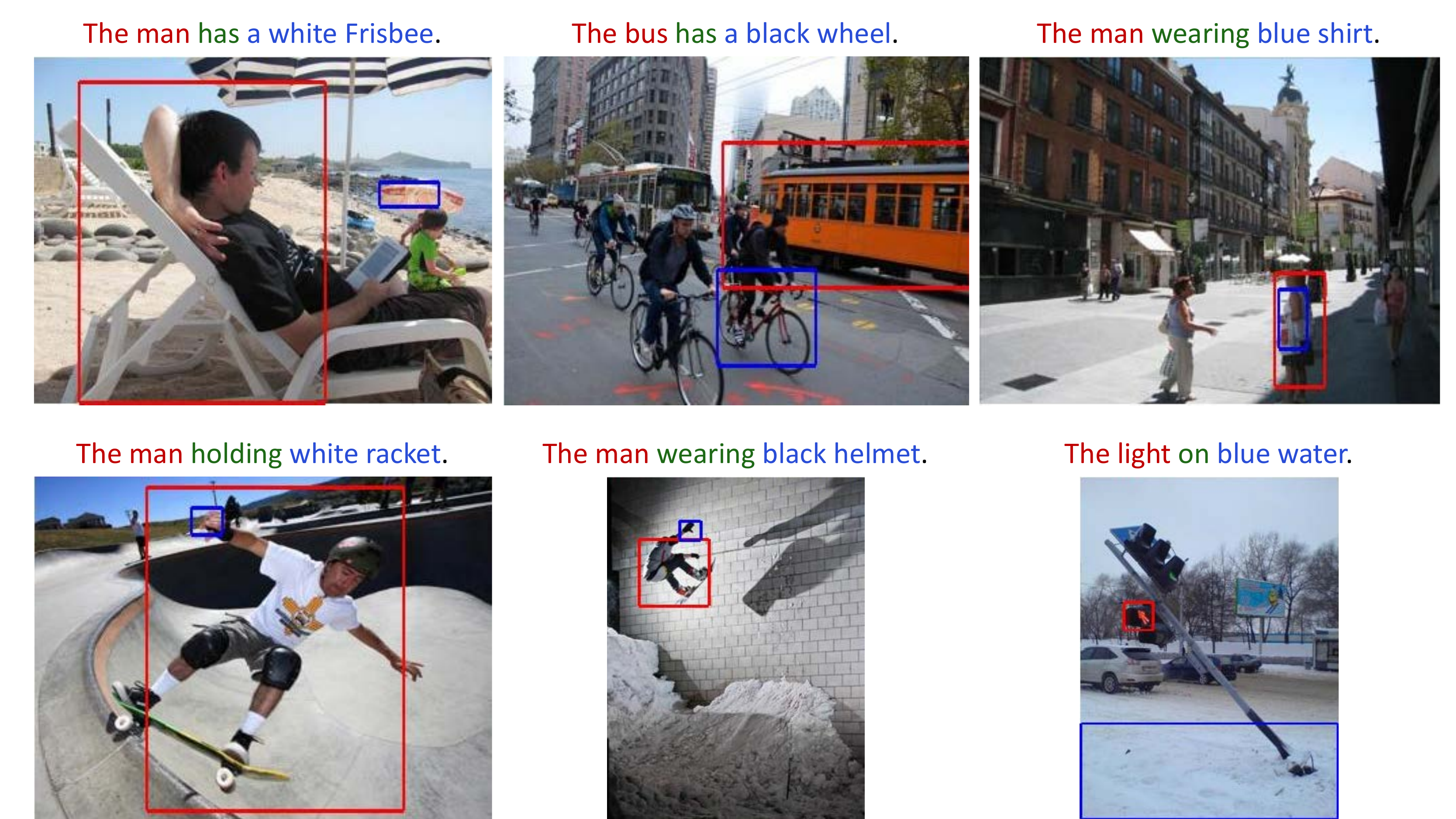}
   \vspace{-4mm}
   \caption{Failure cases of our model. The reasons for failure cases are often due to
   visual ambiguity and illumination.
   {\texttt{subj}-\texttt{pred}-\texttt{obj} are color-coded by red, green, and blue colors according to the output of the POS predictor, respectively.}}
\label{fig:failure}
\end{figure}

%-----------------------4.6--------------------------------------------------
\subsection{Comparison with VRD Model}
In order to demonstrate the flexibility of our model's output,
% the output our model generates, 
\ie, natural language based sentences, we qualitatively compare our model with one of the benchmark models of visual relationship detection (VRD) task. 
We test the VRD benchmark model~\cite{lu2016visual} and our \texttt{MTTSNet} (and with \texttt{+REM}).
The comparison is shown in \Fref{fig:VRD_comparison}. 
%\djkim{Based on the output of POS predictor, we color the words of caption as (red, green, blue) for (\texttt{subj}-\texttt{pred}-\texttt{obj}) respectively. One can see that the predicted POS is correctly aligned with the words in the generated captions.}
While the output of the VRD model is limited to the \texttt{subj-pred-obj} triplet with a smaller number of classes in a closed set, the output of our model has
more flexibility and can contain more contextual information {by virtue of being free form}.
For example, given the same object ``person,'' our model is able to distinguish the fine-grained category, \ie, man and woman.
In addition, our model can provide rich information about the object (\eg, smiling, gray) by virtue of leveraging attribute information of our relational captioning data.
Thus, our relational captioning framework enables higher level interpretation of the objects compared to the relationship detection framework.
% \djkim{By virtue of natural language handling in our captioning task, our model can generate diverse expressions including attributes, which is not covered by the VRD task using the restricted form. 
% Thus, for a fair comparison, we did not include this comparison, because our work deals with an open set problem. }

{Since the output of the VRD task has {a} relatively simple form (\ie, \texttt{subj-pred-obj} triplet) compared to that {of} our captioning framework (caption with free-form and variable length), a VRD model is easier to train given the same relationship detection dataset.
Thus, a direct comparison with a VRD model on the VRD dataset \cite{lu2016visual} is unfair for our method.}
Despite this, we perform quantitative comparisons with VRD models by restricting {the output vocabulary of our model such that  the words {appeared} in the VRD dataset without attributes are only used.}
We use the VRD dataset that contains in total 5000 images with 4000/1000 splits for train/test sets respectively.
%They are labeled with various subject-object bounding box pairs and their corresponding relationship expressions.
Similar to the construction process of our relational captioning dataset, we tokenize the form of triplet expression, \ie, \texttt{subj-pred-obj}, to form natural language expressions, and for each word, we assign the POS class from the triplet association.
By tokenizing, we obtain 160 vocabularies for the VRD dataset.

{We evaluate on this regime in Tables~\ref{table:ablation_VRD} and \ref{table:VRDmetric} with the relational captioning metrics and VRD metrics, respectively. Firstly,}
\Tref{table:ablation_VRD} shows {the comparisons with VRD models~\cite{lu2016visual,yang2018shuffle} on the VRD dataset along with the ablation study.}
Overall, the ablation study shows similar {trends} to that of using our relational captioning dataset {(\cf, \Tref{table:captioning})}. 
{Our \texttt{TSNet} and \texttt{MTTSNet} ({both with and without \texttt{+REM}}) show top performance among the relational captioning models, {of which difference is with and without POS prediction (w/MTL), respectively}. 
This suggests that, even on the VRD dataset, the triplet-stream component is still a strong baseline over others.}
Moreover, interestingly, while the POS classification appears to be an easy and basic task, adding the POS classification in the form of multi-task learning consistently helps the caption generation performance by a noticeable margin in our context, as shown in Tables \ref{table:captioning} and \ref{table:ablation_VRD}.

In the last row, we show the performance of the VRD models by Lu \etal~\cite{lu2016visual} and Yang \etal~\cite{yang2018shuffle} with the relational captioning metrics.
Note that {these VRD models are}
% this VRD model is
designed specifically for triplet classification on the VRD dataset.
Thus, in terms of mAP, 
% {which measures the accuracy of the classification,}
it has an advantage compared to the results of the other relational captioning baselines. 
{Nonetheless, compared to the VRD model, {our} relational captioners (especially our \texttt{MTTSNet+REM}) show favorable performance on Img-Lv Recall and METEOR with a notable margin.
This suggests that the proposed relational captioning framework is advantageous in generating \emph{diverse} and \emph{semantically natural} expressions.
On the other hand,} VRD models are disadvantageous in these aspects because they use {a} closed vocabulary set and predict object classes individually without considering the context.

\Tref{table:VRDmetric} shows the comparison between our \texttt{MTTSNet} ({both with and without \texttt{+REM}}) and other VRD models 
measured on {the} VRD metrics.
Due to the difference of our output type to that of VRD, we use METEOR score thresholds proposed by \cite{johnson2016densecap} as the matching criteria {between model outputs and ground truth labels}. 
{Among the three VRD tasks (\emph{predicate classification}, \emph{phrase detection} and \emph{relationship detection}) defined in~\cite{lu2016visual}, we do not measure \emph{predicate classification} because a simple classification is out of scope for our model,} {but context understanding.}
As shown in the table, our model shows favorable {or comparable} performance to {the VRD} models despite {the fact} that they are specifically designed for the VRD task.
{Also, note that most of the VRD methods take an advantage of strong backbone networks such as ResNet over our \texttt{MTTSNet+REM} that uses VGG-16.
According to the table, our method with the ResNet-50 backbone performs better than all the other competing VRD methods.
}
% \djkim{Also, as we change the backbone architecture from VGG-16 to ResNet, which is broadly used in VRD methods, our model shows similar performance to that of more recent VRD methods.}
This is worth noting in that, as opposed to VRD, our output label space is more complex than that of VRD due to variable caption length {and {a much larger number} 
% diversity
of vocabulary}.
%This is an advantage of being free form, and we believe that this potential aspect could add contribution to VRD models.
% In conclusion, our \texttt{MTTSNet} is comparable with other models using VRD data.

%-----------------------4.7--------------------------------------------------
\subsection{Additional Analysis}

\noindent\textbf{Vocabulary statistics.}
In addition, we measure the vocabulary statistics and compare those of
% them  among 
the frameworks in \Tref{table:diversity}.
The types of statistics measured are:
% 1) the total number of unique words that have been used for the whole test set, 
1) an average number of unique words that have been used to describe an image, and
2) an average number of words to describe each box.
% More 
Specifically, we count the number of unique words in all the predicted sentences and present the average number per image or box.
Thus, the metric {is proportional to} the amount of information we can obtain given an image or a fixed number of boxes.
% The comparison is depicted in \Tref{table:diversity}.
These statistics increase in the order of \emph{Image Cap.}, \emph{Scene Graph}, \emph{Dense Cap.}, and \emph{Relational Cap} ({both with and without \texttt{+REM}}).
{In conclusion, the proposed relational captioning is favorable
% advantageous
in diversity and amount of information ({especially when {the} REM module is added}), compared to both of the traditional object-centric scene understanding frameworks, \ie, \emph{Dense Cap.} and \emph{Scene Graph}.}

\noindent\textbf{Importance transition along the triple-LSTMs.}
{Since we have the three state LSTMs to predict a single word, it might be questionable whether each LSTM learns their own semantic roles properly.
To see the behavior of each LSTMs,}
we visualize the weight transition from each LSTM for each time step. 
For this, given a set of features fed to the triple-stream LSTMs, we compute the L2 norm of the LSTM hidden state vector for each time step {as a measure of importance value}. 
These {values} from the three LSTMs are normalized across time through mean value subtraction.
% and normalized through LSTMs with softmax operation. 
These {values} can be regarded as information or importance quantities.
\Fref{fig:weighttransfer} shows the transitions of the representative values across time. 
As the {POS phase} changes through subject-predicate-object, the weight of the subject LSTM consistently decreases {while that of the object LSTM increases.}
{The} predicate LSTM has {a relatively consistent} intensity between subject and object LSTMs {as the POS changes}.
Thus, LSTMs plausibly disentangle their own roles according to POS.
{In other words, each word in the captions comes from the corresponding LSTM, \eg,~a subject word is generated from the subj-LSTM.}
% Therefore, we can observe the {information transition change in weights or information usage} {as POS changes} from subject to object.}

%\noindent\textbf{Knowledge transfer result for VQA?}
%\djkim{one or two more application will go here.}

\noindent\textbf{Discussion of the failure cases.}
\Fref{fig:failure} shows failure cases of {our} relational captioning.
%\djkim{Based on the output of POS predictor, we color the words of caption as (red, green, blue) for (\texttt{subj}-\texttt{pred}-\texttt{obj}) respectively.}
% \djkim{Note that the predicted POS is still correctly aligned with the words in the generated captions despite the errors in the captions.}
%Similar to \Fref{fig:multibox}, the words in the caption are colored based on the predicted POS from the model.
The captions generated from our method can be inaccurate for several reasons.
One of the important factors is visual ambiguity. 
Ambiguity may come from visually similar but different objects (first column) or by geometric ambiguity (second column). 
{Lastly}, due to illumination, the model may describe the object {with} 
% as
a different color (\eg, ``blue'') (third column).
% In addition, due to dataset bias, the model may predict words that are frequent in the dataset (\eg, ``window'').
% Finally, the model may be weak for predicting unfamiliar objects.
%because of data bias to frequent sentences such as ``building has wall'' or ``window on building''. In addition, false prediction can be caused by similar appearance (\eg, objects that can be confused with ``window'' or ``monitor''), fail in detecting fine-grained detail (\eg, ``man'' and ``woman''). 
{Each of cases requires challenging capabilities, such as geometric reasoning, high resolution spatial representation learning, illumination invariance, \etc, which are all fundamental computer vision challenges.}
we postulate that these problems may be resolved by improving visual feature representation; we leave these failure cases as a future direction.
%{Despite this, 
Note that 
the predicted POS is still correctly aligned with the words in the generated captions.
%}

%\noindent\textbf{Diverse examples of generated captions.}
%We provide examples generated by our model in \Tref{table:diverse_caption} that shows the diversity of the predicates for the same subject-object pairs. 
%Even though we leverage existing relationship datasets, we {extensively modify them and} construct diverse and plausible {augmented} datasets by associating attributes.{}
%{Details are explained in Section 2.2.}

%%%%%%%%%----5----%%%%%%%%%%%%%%%%%%%%%%%%%%%%%%%%%%%%%%%%%%%%%%%%%%%%%%%%%%%%%%%%%%%%%%%%%%%%%
%\section{Supplementary}
%\input{sec_supplementary_CVPR}

% \vspace{-2mm}
\section{Conclusion}
% \vspace{-2mm}
We introduce relational captioning, a new notion which requires a model to localize regions of an image and describe each of the relational region pairs with {natural language}. 
To this end, we propose the MTTSNet, which facilitates POS aware relational captioning. 
In several sibling-tasks, we empirically demonstrate the effectiveness of our framework over scene graph generation and the traditional captioning frameworks.

{Furthermore, our relational captioning can provide dense, diverse, abundant, high-level and interpretable representations in {a} caption form, which is a new way to represent imagery. This allows us to take several advantage over the existing tasks of VRD and \{image, dense\} captionings.
Compared to VRD, our relational captioning deals with ``openset'' (or a much larger set) expression.
The VRD task is restricted to \texttt{subj-pred-obj} combinations, of which term represents a fixed number of classes.\footnote{Suppose the number of classes of each term, \texttt{subj}, \texttt{pred}, and \texttt{obj}, is all same as $C$ in the VRD task. Then, the number of all possible combination of VRD is limited to $C^3$.}
However, the natural language representation we use has free-form with varying lengths, which can represent uncountably many possibilities. 
In addition, as an object can be referred to with expressive and distinctive attributes, which cannot be done in the VRD task, \eg, a man vs. a boy wearing red hat, our output representation is more general. 
%Compared to image-level or dense captioning, our relational captioning is ``denser.''
Finally, as we have shown in our results, our captioning has much higher recall than the other tasks, which can transfer sufficient information to subsequent algorithms and applications.}
In this regard, our work may open {new} interesting applications.
% , \eg, natural language based video summarization~\cite{choi2018contextually} may be benefited by our rich representation.

% \clearpage
\section*{Appendix -- Attribute Enrichment}
\begin{figure}[t]
	\centering 
    %\raisebox{\dimexpr\ht\imagebox-\height}
    \includegraphics[width=1\linewidth,keepaspectratio]{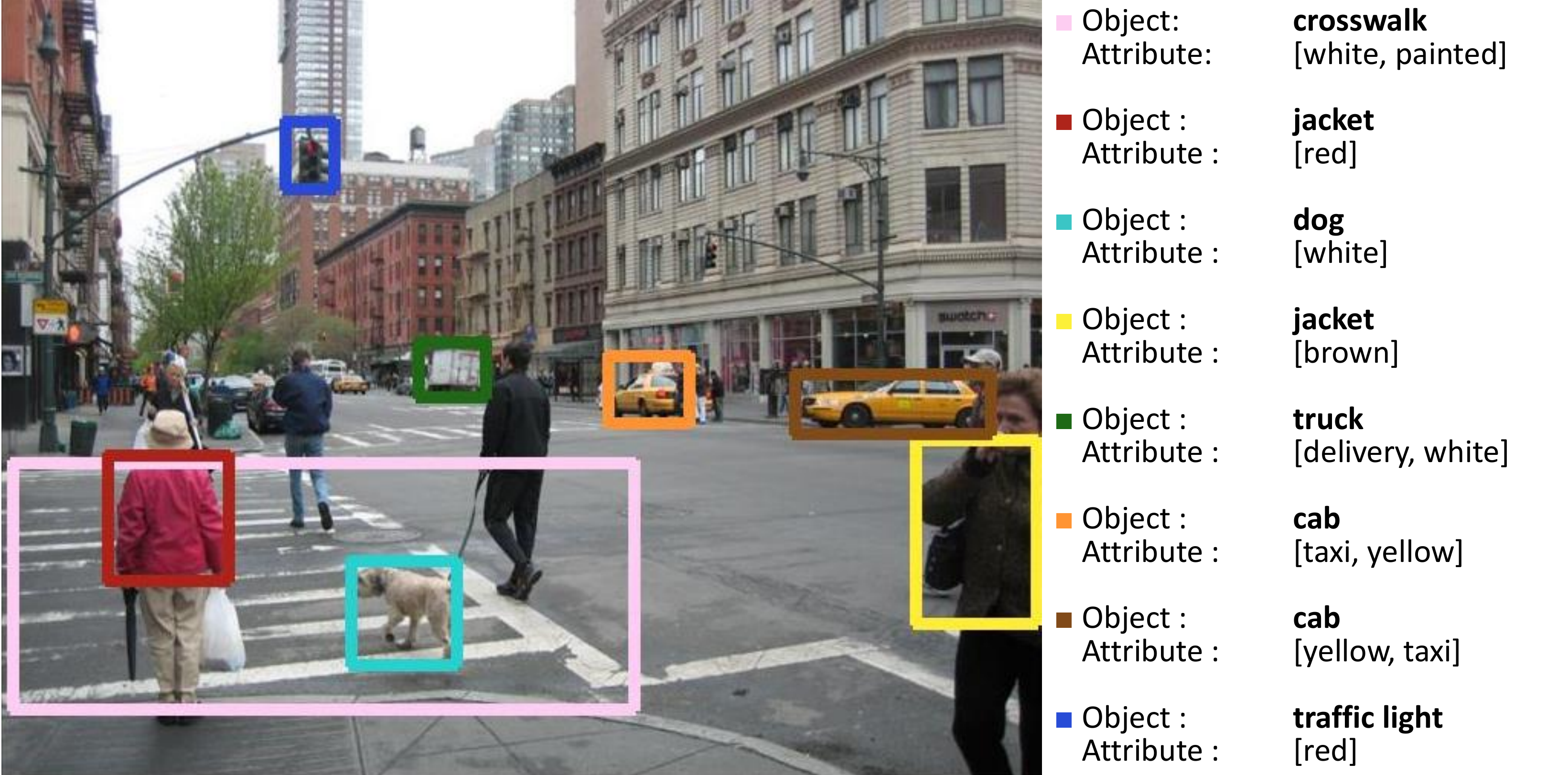}
    \vspace{-6mm}
	\caption{{A sample of the VG attributes dataset.} 
	Each bounding box is labeled with {an} object name and attributes (Attribute labels for a bounding box can be multiple).
    \vspace{-4mm}}
	\label{fig:attributes}
\end{figure}

% \subsection{Enrichment by Attributes}
{As described in \Sref{sec:exp_setup}, we construct the relational captioning dataset based on the VG relationship dataset, but the dataset lacks attribute information in the captions.
To compensate the lack of attributes, we leverage {VG \emph{attribute} dataset}~\cite{krishna2017visual}.
}
% \djkim{
% When constructing relational captioning dataset by utilizing , %it shows limited diversity {of} the words used.
%In order to examine the diverse expression of our relational captioner, we make our relational captioning dataset to have more natural sentences with richer expressions.
%Through observation, 
% {we noticed that the relationship dataset labels lack \emph{attributes} describing the subject and object, which are perhaps what enriches the sentences the most}.
% Therefore, we utilize the {VG \emph{attribute} dataset}~\cite{krishna2017visual}.
% }

{The configuration of the VG attribute dataset is depicted in \Fref{fig:attributes}.
In the dataset, each object bounding box in an image is associated with ``object name'' and ``attributes'' of the object.
Note that each object can have multiple attributes at the same time.
Since the VG relationship dataset and the attribute dataset share the same image set, while \emph{the ground-truth bounding boxes are not shared}, to associate the attribute with our captions, we conduct the process to find corresponding bounding boxes between datasets.
}

% The procedure of the enrichment is as follows.
%Specifically, 
We simply find the attribute that matches the subject/object of the relationship label and {assign}
% attach
it to the subject/object caption label.
In particular, if an attribute label describes the same subject/object for a relationship label while an associated bounding box overlaps enough, the label is considered to be \emph{matched} to the subject/object in the relationship label.

The specific procedure to decide association are as follows:
\begin{enumerate}[leftmargin=4mm]
    \item The {category words} of the subject / object in
    the relationship label and the object names of the attribute label 
    must match, and the boxes should sufficiently overlap (higher IOU than 0.7), 
    \item Among the several boxes satisfying this condition, the box with the highest IOU is selected. 
    \item If a single box is associated with multiple attribute labels, we check the part-of-speech (POS) of candidate attribute labels using the NLTK POS tagger~\cite{loper2002nltk}.
    The words classified as \texttt{(NN, VBN, VBG, VBD, JJ)} are regarded as appropriate candidates for natural attributes. 
    {We filter out the other categories.}
    \item Among the attribute candidates, the words in the original relationship triplet (\emph{i.e.} \texttt{subj-pred-obj}) are excluded from the candidates {to prevent redundancy}.
    \item If there are still {more than one} candidate attributes satisfying all these conditions, we randomly select one among the candidates.
    \item If a subject does not have any matched attribute, ``the'' is added. 
\end{enumerate}
% \vspace{-4mm}
 %\oh{QUESTION: multiple independent attributes from respective different labels? or a single complex attribute consisting of multi-words?}
% In the same box however, there may be several attribute labels.
% First, 
%\oh{COMMENT: move this paragraphs to an appendix section in the end of the paper. 
% what do you prefer? BTW, does TPAMI allow appendix section? please check}\djkim{(It seems any kind of materials can be added. 1 manuscript + maximum 4 extra files (including summary of changes, supplementary material, appendix, previous versions )
%}

{Note that the VG relationship dataset and the VG attribute dataset share the same image set; thus, there exists object-level correspondences. Since we leverage the correspondences, thus, our dataset is likely to follow real distribution of image-description contents in the datasets.
}

\noindent{\textbf{Acknowledgements.}}
This work was supported by the Institute for Information \& Communications Technology Promotion (IITP) (2017-0-01772) grant funded by the Korea government.
T.-H. Oh is partially supported by the IITP grants funded by the Korea government (MSIT) (No. 2019-0-01906, Artificial Intelligence Graduate School Program (POSTECH); No.2021-0-02068, Artificial Intelligence Innovation Hub).

\vspace{0mm}

{\small
\bibliographystyle{ieee}
\bibliography{egbib}
}

% \begin{IEEEbiography}[{\includegraphics[width=1in,height=1.25in,clip,keepaspectratio]{mshell}}]{Michael Shell}

\begin{IEEEbiography}[{\includegraphics[width=1in,height=1.25in,clip,keepaspectratio]{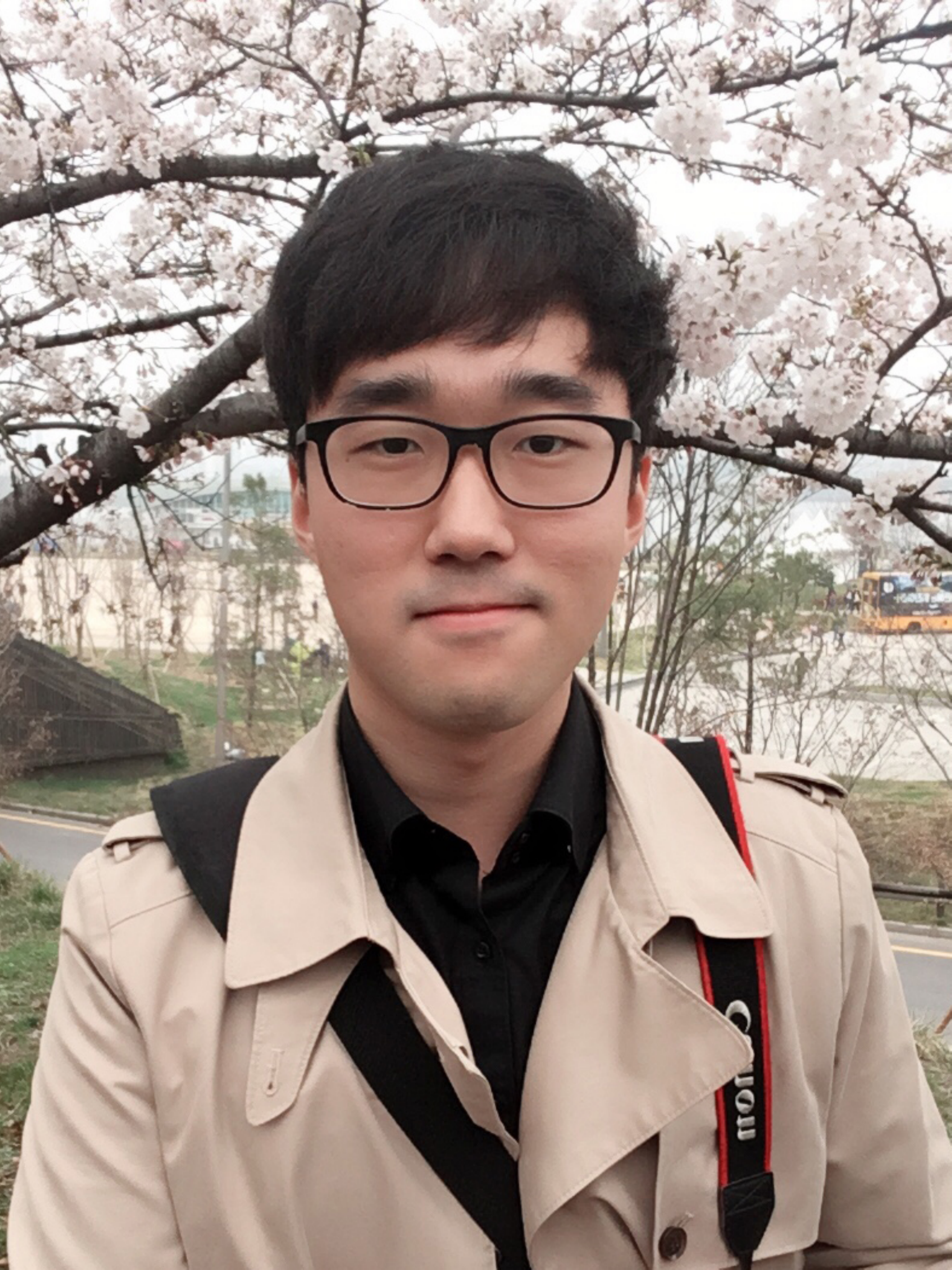}}]{Dong-Jin Kim}
received the B.S. degree, M.S. degree, and Ph.D. degree in Electrical Engineering from Korea Advanced Institute of Science and Technology (KAIST), Daejeon, South Korea, in 2015, 2017, and 2021, respectively.
%He is currently working towards the Ph.D. degree in Electrical Engineering at KAIST.
He was a research intern in the Visual Computing Group, Microsoft Research Asia (MSRA).
He was awarded a silver prize from Samsung Humantech paper awards and Qualcomm Innovation awards.
His research interests include high-level computer vision such as language and vision and human behavior understanding. 
He is a student member of the IEEE.
\end{IEEEbiography}

\begin{IEEEbiography}[{\includegraphics[width=1in,height=1.25in,clip,keepaspectratio]{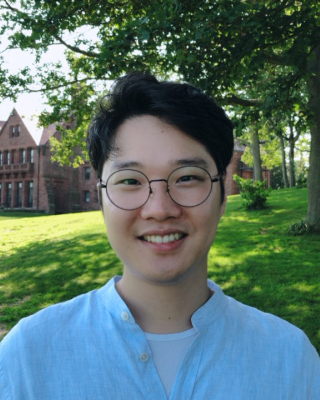}}]{Tae-Hyun Oh}
is an assistant professor with Electrical Engineering (adjunct with Graduate School of AI and Dept. of Creatiative IT Convergence) at POSTECH, South Korea. He is also a research director at OpenLab, POSCO-RIST, South Korea. 
He received the B.E. degree (First class honors) in Computer Engineering from Kwang-Woon University, South Korea in 2010, and the M.S. and Ph.D. degrees in Electrical Engineering from KAIST, South Korea in 2012 and 2017, respectively.
Before joining POSTECH, he was a postdoctoral associate at MIT CSAIL, Cambridge, MA, US, and was with Facebook AI Research, Cambridge, MA, US. 
He was a research intern at Microsoft Research in 2014 and 2016. He serves as an associate editor for the Visual Computer journal.
He was a recipient of Microsoft Research Asia fellowship, Samsung HumanTech thesis gold award, Qualcomm Innovation awards, top research achievement awards from KAIST, and CVPR'20 outstanding reviewer award. 
\end{IEEEbiography}

\begin{IEEEbiography}[{\includegraphics[width=1in,height=1.25in,clip,keepaspectratio]{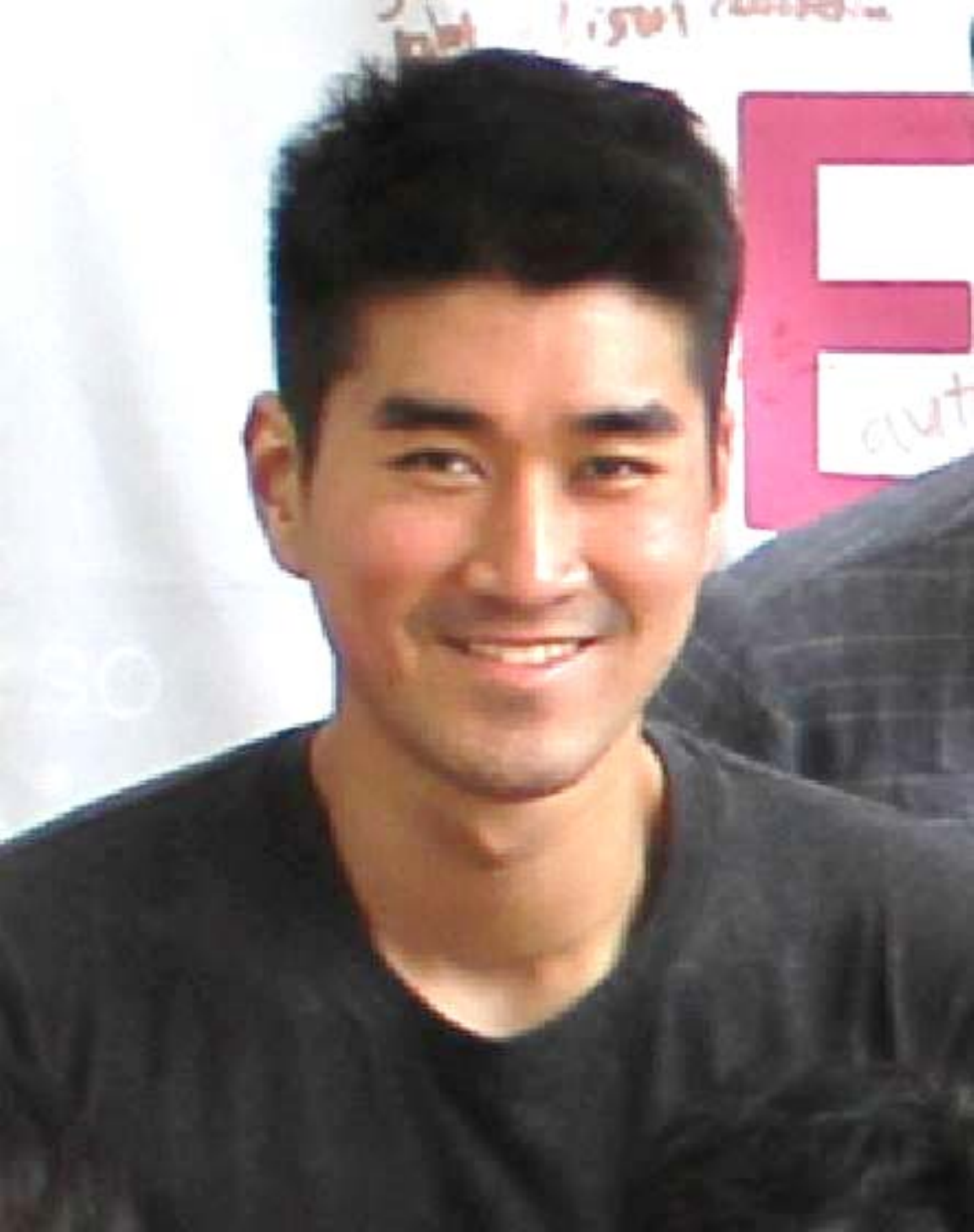}}]{Jinsoo Choi}
received his B.S., M.S., and Ph.D. degrees in Electrical Engineering from Korea Advanced Institute of Science and Technology (KAIST) in 2013, 2015, and 2020 respectively.
He is an incoming machine learning video engineer at Apple.
He received the grand prize from the Electronic Times paper awards hosted by the Ministry of Science and ICT, Rep. of Korea, silver prize from Samsung Electro-Mechanics paper awards, silver prize from Samsung Humantech paper awards, Qualcomm  Innovation  awards, recognition as top research achievements and top 1\% research achievements from KAIST annual and biannual R\&D reports.
His research interests include deep learning, computer vision, and computer graphics with an emphasis on video enhancement and processing.
\end{IEEEbiography}

\begin{IEEEbiography}[{\includegraphics[width=1in,height=1.25in,clip,keepaspectratio]{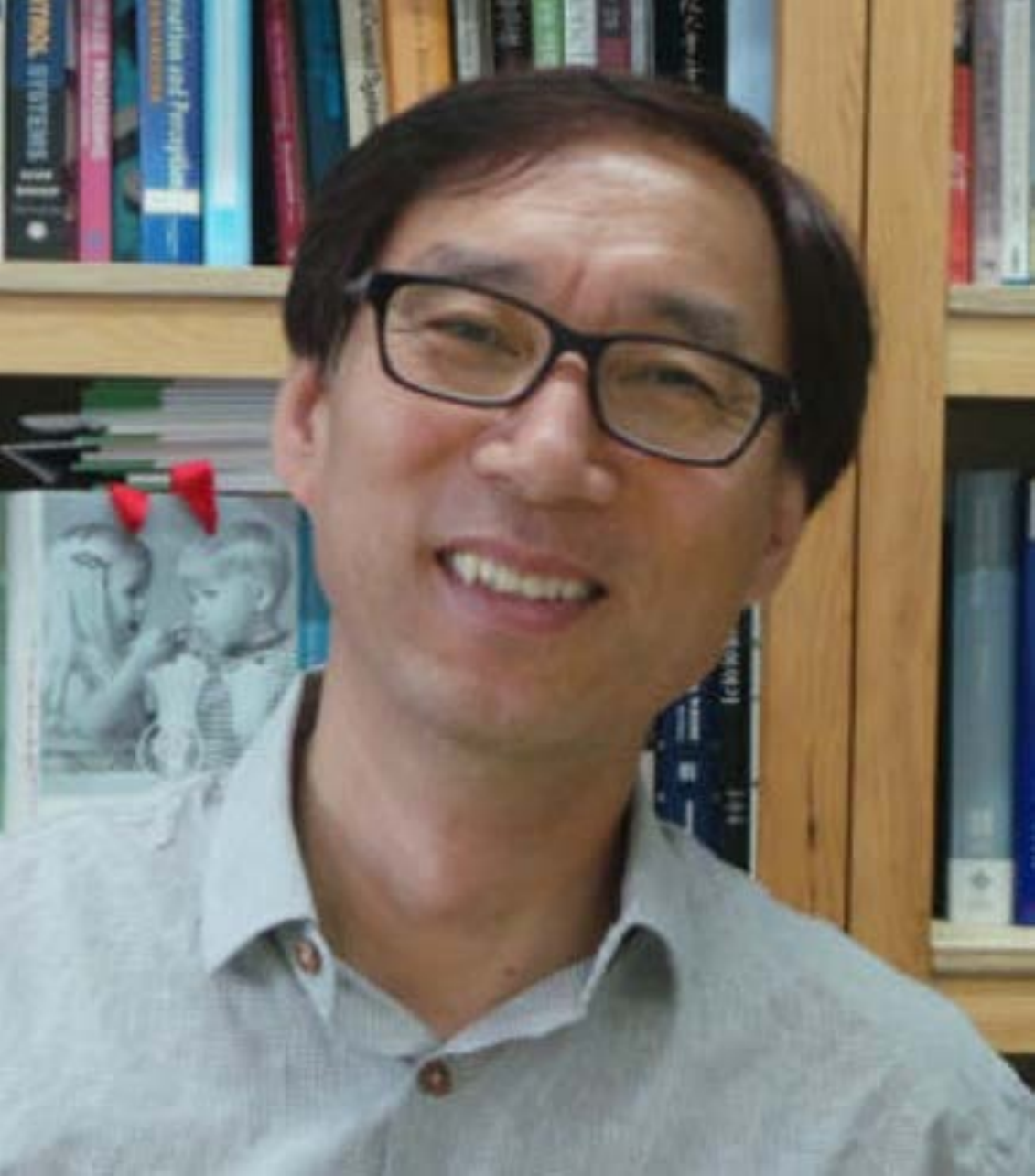}}]{In So Kweon}
received the BS and MS degrees in mechanical design and production engineering from Seoul National University, Seoul, South Korea, in 1981 and 1983, respectively, and the PhD degree in robotics from the Robotics Institute, Carnegie Mellon University, Pittsburgh, Pennsylvania, in 1990. He is an professor with the Electrical Engineering Department, KAIST, South Korea. He worked for the Toshiba R\&D Center, Japan, and joined the Department of Automation and Design Engineering, KAIST, Seoul, South Korea, in 1992, where he is currently a professor with the Department of Electrical Engineering. He is a recipient of the Best Student Paper Runner-up Award at the IEEE Conference on Computer Vision and Pattern Recognition (CVPR 09). His research interests include camera and 3D sensor fusion, color modeling and analysis, visual tracking, and visual SLAM. He was the program co-chair for the Asian Conference on Computer Vision (ACCV 07) and was the general chair for the ACCV 12. He is also on the editorial board of the International Journal of Computer Vision. He is a member of the IEEE
and KROS.
\end{IEEEbiography}

% \clearpage
% \section{contents for response letter}
% \input{sec_response_letter2}

% \clearpage
% \section{contents for cover letter}
% \input{sec_cover_letter2}

\end{document}